\documentclass[pdflatex,sn-aps]{sn-jnl}

\usepackage{graphicx}%
\usepackage{multirow}%
\usepackage{amsmath,amssymb,amsfonts}%
\usepackage{amsthm}%
\usepackage{mathrsfs}%
\usepackage[title]{appendix}%
\usepackage{xcolor}%
\usepackage{textcomp}%
\usepackage{manyfoot}%
\usepackage{booktabs}%
\usepackage{algorithm}%
\usepackage{algorithmicx}%
\usepackage{algpseudocode}%
\usepackage{listings}%

\usepackage[caption=false,font=footnotesize]{subfig}

\newif\ifreview
\reviewfalse

\newif\ifpreprint
\preprintfalse

\newif\iflong
\longfalse

\begin{document}

\title{FREQuency ATTribution: Benchmarking Frequency-based Occlusion for Time Series Data}

\author*[1]{\fnm{Dominique} \sur{Mercier}}\email{dominique.mercier@dfki.de}
\author[1]{\fnm{Andreas} \sur{Dengel}}\email{andreas.dengel@dfki.de}
\author[1]{\fnm{Sheraz} \sur{Ahmed}}\email{sheraz.ahmed@dfki.de}
\affil[1]{\orgdiv{Smart Data \& Knowledge Services}, \orgname{DFKI GmbH}, \orgaddress{\street{Trippstadter Str. 122}, \city{Kaiserslautern}, \postcode{67663}, \state{RLP}, \country{Germany}}}

\abstract{Deep neural networks are among the most successful algorithms in terms of performance and scalability across different domains. However, since these networks are black boxes, their usability is severely restricted due to a lack of interpretability. Existing interpretability methods do not address the analysis of time-series-based networks specifically enough. This paper shows that an analysis in the frequency domain can not only highlight relevant areas in the input signal better than existing methods but is also more robust to fluctuations in the signal. In this paper, \textit{FreqAtt} is presented — a framework that enables post-hoc interpretation of time-series analysis. To achieve this, the relevant frequencies are evaluated, and the signal is either filtered or the relevant input data is marked. \textit{FreqAtt} is evaluated using a wide range of statistical metrics to provide a broad overview of its performance. The results show that using frequency-based attribution, especially in combination with traditional attribution on top of the frequency-optimized signal, provides strong performance across different metrics.}

\keywords{Deep Learning, Time Series, Attribution, Fast Fourier Transformation, Artificial Intelligence}

\maketitle

\section{Introduction}
\label{sec:introduction}
Deep neural networks have become one of the standard solutions for processing large datasets. Recently, they have achieved outstanding results across various application areas, including regression and forecasting tasks in addition to the commonly used classification tasks. However, despite their excellent performance, neural networks cannot be applied in all domains, as they are black-box approaches~\cite{bibal2021legal}. Especially for safety-critical applications, it is essential to provide an explanation for the predictions made by such algorithms. This serves as a basis for validating the correctness of the prediction. In addition, interpretability has become essential due to the GDPR~\cite{zaeem2020effect}. As a result, a research community has emerged that focuses on the interpretability of neural networks — namely, eXplainable Artificial Intelligence (XAI)~\cite{arrieta2020explainable}.

The XAI community has developed different methods to explain the prediction of neural networks, including intrinsic and post-hoc methods~\cite{burkart2021survey}. While both categories offer different advantages, an intrinsic explanation is characterized by the inherently explainable reasoning process of the network. This kind of interpretability, however, limits the architecture of the network, which restricts not only the performance but also the flexibility of the approach. For example, it is not possible to adapt every state-of-the-art architecture to make the prediction process interpretable. In contrast, post-hoc methods are mostly independent of the network architecture, as they only work with the prediction of the network. The most frequently used post-hoc methods are attribution methods, which highlight relevant input values. The class of attribution methods can again be divided into gradient-based and permutation-based approaches~\cite{mercier2022time}. Gradient-based approaches usually provide faster results, are more closely tied to the network parameters, but are noisier and require more cognitive effort. These algorithms use the backpropagation algorithm employed in the training phase to identify which input values have the largest influence on the output~\cite{rumelhart1986learning}. Permutation-based approaches, on the other hand, require more runtime but present the results in a smoother form, demanding less cognitive effort. These approaches slightly change the input and measure the impact on the output of the network, making them independent of the architecture.

A problem that is not addressed by any of the existing approaches is the interaction between data points. In time-series analysis, it is possible that different data points influence each other; this interaction can even occur across signal channels~\cite{mercier2022timereise}. In gradient-based methods, this problem is not considered since they rely solely on the optimization of the network. Through backpropagation, the network is optimized to achieve the best result, but this can lead to special properties of the network being exploited, resulting in a distorted statement regarding interpretability~\cite{adebayo2018sanity}. Permutation-based methods mostly rely on local proximity~\cite{zeiler2013visualizing}. Although there are methods that consider the local independence of data points, these are rare. One reason for this is that these methods were originally developed for image analysis, where the local dependencies of pixels and the interactions between channels are firmly defined. To address this problem, it is necessary to transform the time series into frequency space. Attribution in frequency space avoids the problem of temporal components, since a change in the frequencies present in the signal affects the entire signal. This further allows the use of permutation-based attribution methods, which optimize the signal so that the relevant frequencies are highlighted, while less relevant frequencies can be filtered out. Another advantage of this approach is its flexibility, as it is independent of the architecture and can be used with any network.

\section{Related Work}
\label{sec:related}
The field of interpretability methods includes many approaches that can be applied across a wide variety of domains. However, most of these methods were initially developed for image analysis and must therefore be adapted to other domains accordingly. Arrieta et al. provide a good overview of existing interpretability methods~\cite{arrieta2020explainable}. Regardless of the modality, the goal of each of these methods is to make the reasoning process of neural networks more explainable, enabling their use in safety-critical applications. This is currently not possible due to ethical and legal limitations that apply to black-box applications.

\subsection{Feature Attribution}
To solve this problem, there are two different approaches: the intrinsic and the post-hoc approach. Both approaches, as well as their corresponding technical implementations, are described in~\cite{burkart2021survey}. In principle, the use of intrinsic methods limits the choice of architecture, making it impossible to use this approach once the network has already been trained.

In contrast, post-hoc methods — in particular, attribution methods — offer the possibility of subsequently adding interpretability. A comprehensive overview of these methods and their advantages and disadvantages can be found in~\cite{mercier2022time}. However, since these methods were initially developed for image analysis, they encounter several challenges when applied to time-series analysis. 

The category of attribution methods can generally be divided into two subcategories. The first of these, the gradient-based approaches, use the backpropagation algorithm, which is also used to train neural networks~\cite{simonyan2013deep}. Based on the internal network parameters, it is possible to determine which input values have influenced the prediction. One key advantage of these approaches is that they do not require any forward passes and can therefore be calculated quickly. In contrast, permutation-based approaches use the forward pass to calculate the influence of changes in the input signal on the output value. This means that these methods do not have direct access to the internal values of the network but are slower to compute due to the required forward passes. In~\cite{nielsen2022robust}, it was shown that both categories can be used, but in different contexts.

Existing attribution methods to explain model predictions at the feature level and form the foundation on which frequency-based attribution builds cover and is compared against are: \textit{Saliency}~\cite{zeiler2013visualizing} maps which compute the gradient of the model output with respect to the input, assigning higher relevance to those input dimensions where small perturbations induce large changes in the prediction, and are often used as a baseline for more advanced explanation techniques. \textit{Integrated Gradients}~\cite{sundararajan2017axiomatic} which extend this idea by accumulating gradients along a path from a baseline to the input, which reduces gradient saturation effects and enforces desirable axioms such as sensitivity and implementation invariance, making the resulting attributions more stable. Model-agnostic techniques like \textit{KernelSHAP}~\cite{lundberg2017unified}, \textit{LIME}~\cite{ribeiro2016should}, and \textit{Occlusion}~\cite{zeiler2013visualizing} complement gradient-based methods and are also relevant when interpreting models in the frequency domain. \textit{KernelSHAP} approximates Shapley values via a weighted linear regression in a locally sampled neighborhood, providing theoretically grounded feature attributions that can, in principle, be applied to frequency coefficients instead of raw features. \textit{LIME} fits a simple interpretable surrogate model around a given instance by perturbing inputs and weighting them by proximity, and the resulting local coefficients can similarly be used to interpret the influence of frequency-domain representations of the input.

Because the different attribution methods vary so widely, it is necessary to use a broad set of metrics to evaluate them. In particular, ~\cite{mercier2022time} has shown that, depending on the chosen metric, different groups of attribution methods are favored. In addition to performance and the insertion and deletion tests, runtime, \textit{Continuity}, \textit{Sensitivity}, and \textit{Infidelity} have emerged as relevant metrics in the literature. Details on the individual metrics can be found in~\cite{yeh2019fidelity} and~\cite{ancona2017towards}. However, it has been shown that no single approach performs optimally across all metrics.

Ultimately, there are very few approaches specifically developed for time series that address the properties associated with them. \textit{DynaMask}~\cite{li2023dynamask} is one such approach, in which attribution is optimized based on permutations. A disadvantage of this approach, however, is its extremely long runtime and the binary classification of features into relevant and irrelevant categories.

\subsection{Frequency Importance}
Multiple studies examine the impact of the spectral domain on deep neural networks, revealing how distinct frequency components shape model behavior and performance. In particular, the roles of low- and high-frequency content have garnered significant attention across various works~\cite{abello2021dissecting,fridovich2022spectral,maiya2021frequency,rahaman2019spectral,wang2020high}. Jo and Bengio~\cite{jo2017measuring} were the first to demonstrate that DNNs tend to rely on superficial noise rather than high-level concepts in the dataset as classification evidence. By dissecting the frequency spectrum into disjoint sections and selectively removing them, this approach tests—and robustly confirms across diverse deep learning architectures and datasets—the hypothesis that convolutional neural networks  rely more heavily on high-frequency features imperceptible to humans.

High-frequency biases in neural networks have also often been linked to adversarial attacks~\cite{tsuzuku2019structural,zhou2021high}. In addition to studies on the impact of different frequency ranges, several works examine how models weigh the amplitude and phase components of an image’s frequency spectrum in decision-making. Chen et al.~\cite{chen2021amplitude}, for instance, demonstrate that neural networks predominantly rely on the amplitude spectrum, while humans find features in the phase spectrum more robust; they introduce Amplitude-Phase-Recombination augmentation to enhance model robustness against adversarial attacks, common corruptions, and out-of-distribution detection by compelling networks to prioritize human-aligned phase information.

Wang et al.~\cite{wang2020towards} pioneered frequency-based explanations for DNN image classifiers by measuring the contribution of specific frequency regions through targeted ablation. Their approach primarily elucidates prediction shifts triggered by nearly imperceptible perturbations in mid- to high-frequency ranges. Another related approach by Kasmi et al.~\cite{kasmi2023assessment} focuses on the attribution in the space-scale domain using the wavelet transform.

\section{Evaluated Method}
\label{sec:approach}
This paper introduces \textit{FreqAtt}, a novel frequency permutation-based attribution algorithm for the analysis of time-series-based neural networks, building upon the work of~\cite{Schmeisser2025}. \textit{FreqAtt} transforms the input into the frequency domain and applies an occlusion mask to remove specific frequencies from the input signal. The modified input signal is then evaluated using a forward pass through the already trained network, and the relevance of each frequency is determined based on the deviation in the prediction. This process is performed for different frequencies and combinations of frequencies and the results are aggregated accordingly, allowing an attribution map to be created in frequency space. Furthermore, this approach can be combined with a traditional attribution method to further highlight the relevant parts of the signal. This is achieved by first optimizing the input using frequency-based attribution and then applying a traditional attribution method on top of the optimized signal.

\subsection{Mathematical Formulation}

To understand the differences between traditional \textit{Occlusion}~\cite{zeiler2013visualizing} and frequency occlusion, both are described below. In the case of traditional occlusion, the masks are applied directly in the feature space. In contrast, frequency attribution starts with the input $x$, transforms it into the frequency domain, applies the occlusion masks, and then transforms the result back to the original space to recompute the predictions and the impact of the occlusion. This results in a major difference compared to normal occlusion, as occluding a frequency can influence the entire sequence rather than being limited to a local time position. Capturing a pattern using traditional occlusion can therefore be difficult, as patterns may be distributed over multiple time steps, and the occlusion window size must be adjusted accordingly. In contrast, frequency occlusion can capture such patterns more easily since it is not restricted to local changes. Furthermore, it is more robust to small fluctuations in the signal, as these mostly affect higher frequencies, which can be filtered out. Similarly, identifying patterns is easier, as they often correspond to specific frequencies.

\subsubsection{Traditional Attribution}
In Equation~\ref{eq:occ} illustrates the traditional occlusion mechanism. Let $\Psi_{class}$ be a trained network that projects an input $x \in \mathbb{R}^{t\times s}$ to a space $\mathbb{R}^{c}$ such that $t$ corresponds to the time steps, $s$ to the channels and $c$ to the classes. For the following formulation $s=1$ is assumed, but in theory every value is possible. The cardinality of a set is described using $ |\cdot| $ and the absolute values of a vector are described using $ ||\cdot||$. Furthermore, $\Phi_{occ}(x,i)$ describes the occlusion applied to an input $x$ using the occlusion mask $n^{i}$ such that $n_{i}^{org}$ is set to zero at position $i$ and to one on the remaining positions as shown in Equation~\ref{eq:occ_detail}. The equation describes the case in which the occlusion mask contains a single zero at one position; however, this can be extended so that a window is covered instead of a single position. Equation~\ref{eq:occ} shows that the output of the occlusion applied to an input is multiplied with a vector that is zero at the occluded position and one at every other position. This leads to an attribution map $a_{occ}$ in the feature space, $\mathbb{R}^{t\times s}$ which can be multiplied with the input to attribute the given features of $x$.

\begin{equation}
\label{eq:occ}
a_{occ} \left ( x \right ) = \sum_{i=0}^{\left | x \right | } \Psi_{class}  \left ( \Phi_{occ} \left ( x, i \right ) \circ n_{i}^{org} \right )
\end{equation}

\begin{equation}
    \label{eq:occ_detail}
n_{i}^{org} = \left ( n_{0},...,n_{j},...,n_{\left | x \right |} \right ) \; , \; n_{i} = 0 \; , \; \forall j \neq i : n_{j} = 1
\end{equation}

\subsubsection{Frequency Attribution}
For detailed information about fast fourier transformation (FFT) the reader is referred to Nussbaumer et al.~\cite{nussbaumer1981fast}. The calculation of the frequency attribution $a_{freq}$ shares some properties with the traditional occlusion approach; however, it differs in terms of the output space. Equation~\ref{eq:freq_attr} shows that the occlusion $\Phi_{occ}(x,i)$ is applied to $\Omega_{fft}(x)$ which is defined as the full complex FFT vector. For this purpose, a set of occlusion masks defined in Equation~\ref{eq:freq_attr_detail}, was created similarly to the previous occlusion masks but in frequency space. In this case, the occlusion masks affect a set of frequencies instead of direct input values, leading to a change in the complete signal rather than a local change fixed to a specific time step and channel. It should be noted that during the experiments, the use of only the magnitude spectrum was also evaluated; however, better results were achieved using the full complex FFT vector.

\begin{equation}
\label{eq:freq_attr}
a_{freq} \left ( x \right ) = \sum_{i=0}^{\left | \Omega_{fft} \left ( x \right ) \right | } \Psi_{class}  \left ( \Phi_{occ} \left ( \Omega_{fft} \left ( x \right ) ,i \right ) \circ n_{i}^{freq}\right )
\end{equation}

\begin{equation}
    \label{eq:freq_attr_detail}
n_{i}^{freq} = \left ( n_{0},...,n_{j},...,n_{\left | \Omega_{fft} \left ( x \right ) \right | } \right ) \; , \; n_{i} = 0 \; , \; \forall j \neq i : n_{j} = 1
\end{equation}

To use the frequency attribution, it is necessary to transform it back to the original input space. Equation~\ref{eq:freq_occ} shows how the attribution is applied to produce an attribution in the input space. Here, $a_{occ}$, the attribution of the frequencies, is multiplied with the input $x$ in the frequency domain using $\Omega_{fft}$ and the result is transformed back using the inverse transform $\Omega_{fft}^{-1}$. The transformed output is then subtracted from the original input x $x$, and the absolute values are computed to obtain $a_{freqocc}$, the final attribution map in the feature space.

\begin{equation}
\label{eq:freq_occ}
a_{freqocc} \left (x \right ) = \left \| x - \Omega_{fft}^{-1} \left (\Omega_{fft} \left (x \right ) \circ a_{freq} \left ( x \right ) \right ) \right \|
\end{equation}

In addition, it is possible to apply a mask function to $a_{frea}$ or directly apply it to the input $x$ using $\Omega_{fft}^{-1}$ to produce an optimized version of $x$ aligned to the predicted class that does not include frequencies of other classes. This procedure is shown in Equation~\ref{eq:freq_occ}, where only the subtrahend is used without taking the absolute values.

Lastly, it is possilbe to apply a traditional attribution method on top of the optimized signal to further highlight the relevant parts of the signal. This is shown in Equation~\ref{eq:combined} where $\Upsilon_{attr}$ describes a traditional attribution method such as occlusion, integrated gradients or saliency maps. The benefit of this combined approach is that the signal is already optimized to contain only relevant frequencies, making it easier for the traditional attribution to highlight important regions. This leads to less noisy explanations that require less cognitive effort to interpret.

\begin{equation}
\label{eq:combined}
a_{combined} \left (x \right ) = \Upsilon_{attr} \left ( a_{freqocc} \left (x \right ) \right )
\end{equation}

\subsection{Theoretical Runtime Evaluation}
The runtime is mainly determined by the number of frequencies and the window size of the occlusion applied over the frequencies. Let $w$ denote the window size of the occlusion and $f$ the number of frequencies of an input $x$. Then the resulting runtime is $O(\frac{f}{w})$ but as $w$ is constant, the final runtime is $O(f)$. In contrast to that, the traditional occlusion linearly increases with the sequence length. Therefore, the runtime is $O(t)$ when $t$ corresponds to the number of time steps. Ultimately, it is not possible to determine which approach is faster, as this depends greatly on both the number of frequencies and the length of the time series. However, it can be stated that the length of the time series is irrelevant for frequency occlusion. With respect to the increase in time steps, the scaling of frequency attribution is $O(1)$ if the number of frequencies present in the signal remains constant.

\section{Datasets}
\label{sec:datasets}

\begin{table}[!t]
\footnotesize
\renewcommand{\arraystretch}{1.3}
\caption{\textbf{Datasets} from different domains. The selection covers a broad variety to address different classification problems related to time series analysis.}
\label{tab:datasets}
\centering
\begin{tabular}{l|l|r|r|r|r|r|r}
\textbf{Domains} & \textbf{Dataset} & \textbf{Train} & \textbf{Val} & \textbf{Test} & \textbf{Steps} & \textbf{Chls} & \textbf{Cls} \\
\hline\hline
\multirow{3}{*}{Manufacturing}
& Anomaly~\cite{siddiqui2019tsviz} & 35,000 & 15,000 & 10,000 & 3 & 50 & 2 \\
& ElectricDevices         & 6,249 & 2,677 & 7,711 & 1     & 96    & 7 \\
& FordA                   & 2,521 & 1,080 & 1,320 & 1     & 500   & 2 \\
\hline
\multirow{2}{*}{Transportation}
& AsphaltPavementType     & 739   & 316   & 1,056 & 1     & 2,371 & 3 \\
& AsphaltRegularity       & 526   & 225   & 751   & 1     & 4,201 & 2 \\
\hline
\multirow{5}{*}{Communications}
& CharacterTrajectories   & 996   & 426   & 1,436 & 3     & 182   & 20 \\
& FaceAll                 & 392   & 168   & 1,690 & 1     & 131   & 14 \\
& SpokenArabicDigits      & 4,620 & 1,979 & 2,199 & 13    & 93    & 10 \\
& UWaveGestureLibrary     & 628   & 268   & 3,582 & 1     & 945   & 8 \\
& Wafer                   & 700   & 300   & 6,164 & 1     & 152   & 2 \\
\hline
\multirow{3}{*}{Public Health}
& ECG5000                 & 350   & 150   & 4,500 & 1     & 140   & 5 \\
& MedicalImages           & 267   & 114   & 760   & 1     & 99    & 10 \\
& NonInvasiveFetalECG     & 1,260 & 540   & 1,965 & 1     & 750   & 42 \\
\hline
\multirow{3}{*}{Food \& Agriculture}
& InsectWingbeat          & 17,500 & 7,500 & 25,000 & 200 & 22    & 10 \\
& Strawberry              & 430   & 183   & 370   & 1     & 235   & 2 \\
& SwedishLeaf             & 350   & 150   & 625   & 1     & 128   & 15 \\
\hline
\multirow{4}{*}{Others}
& LSST                    & 1,722 & 737   & 2,466 & 6     & 36    & 14 \\
& ShapesAll               & 420   & 180   & 600   & 1     & 512   & 60 \\
& StarLightCurves         & 700   & 300   & 8,236 & 1     & 1,024 & 3 \\
& TwoPatterns             & 700   & 300   & 4,000 & 1     & 128   & 4 \\
\end{tabular}
\end{table}

The experiments below were conducted across multiple datasets of the UEA \& UCR repository~\cite{tsc2021datasets}. The datasets were selected in a way that covers multiple different aspects such as different number of channels, classes, dataset size and problem statement. The selection is shown in Table~\ref{tab:datasets}. Furthermore, the selection covers different domains to emphasize on the broad applicability of the methods as well as to highlight limitations. This way, the selection of datasets presents a comprehensive set covering both easy and difficult datasets concerning the generation of data. In addition to the UEA \& UCR datasets, a synthetic anomaly detection dataset was used~\cite{siddiqui2019tsviz}. For the detailed visual and numerical analysis the \textit{CharacterTrajectories} dataset was used as it provides a good balance between the complexity and interpretability. Furthermore, this was the only dataset that provided a good visual representation of the time series data in the 2D space.

\section{Experiments \& Results}
\label{sec:experiments}

\begin{table}[!t]
\small
\renewcommand{\arraystretch}{1.3}
\caption{\textbf{Accuracy} of InceptionTime. The table shows the accuracy and F1-scores for all datasets.}
\label{tab:accuracy}
\centering
\begin{tabular}{l|r|r|r}
\textbf{Dataset} & \textbf{Accuracy} & \textbf{F1 Macro} & \textbf{F1 Micro} \\
\hline\hline
Anomaly~\cite{siddiqui2019tsviz}
                        & 0.9877 & 0.9778 & 0.9876 \\
ElectricDevices         & 0.6719 & 0.6077 & 0.6590 \\
FordA                   & 0.9462 & 0.9462 & 0.9462 \\
\hline
AsphaltPavementType     & 0.9290 & 0.9216 & 0.9287 \\
AsphaltRegularity       & 0.8136 & 0.8057 & 0.8063 \\
\hline
CharacterTrajectories   & 0.9721 & 0.9696 & 0.9720 \\
FaceAll                 & 0.7604 & 0.8317 & 0.7512 \\
SpokenArabicDigits      & 0.9927 & 0.9927 & 0.9927 \\
UWaveGestureLibraryAll  & 0.9305 & 0.9296 & 0.9298 \\
Wafer                   & 0.9958 & 0.9890 & 0.9958 \\
\hline
ECG5000                 & 0.9464 & 0.5842 & 0.9379 \\
MedicalImages           & 0.7039 & 0.5590 & 0.6706 \\
NonInvasiveFetalECG     & 0.9506 & 0.9479 & 0.9505 \\
\hline
InsectWingbeat          & 0.7054 & 0.7053 & 0.7053 \\
Strawberry              & 0.8486 & 0.8448 & 0.8518 \\
SwedishLeaf             & 0.8304 & 0.8164 & 0.8129 \\
\hline
LSST                    & 0.2320 & 0.1902 & 0.2614 \\
ShapesAll               & 0.8883 & 0.8840 & 0.8840 \\
StarLightCurves         & 0.9458 & 0.9084 & 0.9410 \\
TwoPatterns             & 1.0000 & 1.0000 & 1.0000 \\
\end{tabular}
\end{table}

To ensure that the results are meaningful, a state-of-the-art model called InceptionTime~\cite{fawaz2020inceptiontime} was trained for each dataset. As the explanation quality depends on the model’s performance, it is crucial to demonstrate that InceptionTime achieves performance sufficient to produce stable explanations. Therefore, Table~\ref{tab:accuracy} presents all the results. To achieve these results, the training data was divided into training and validation sets using a $70$/$30$ split. Additionally, early stopping and a learning rate scheduler were used to obtain the best possible models. The training was limited to a maximum of $50$ epochs with a batch size of $32$. The results demonstrate that the network achieved good classification performance across nearly all datasets. However, for the \textit{ECG5000} and \textit{MedicalImages} datasets, the macro F1 scores show a substantial drop compared to their micro F1 scores. This discrepancy arises due to the severe class imbalance within these datasets. Apart from these two datasets, the macro F1 scores indicate that the model effectively learned representations for all classes. Furthermore, the micro F1 scores for both \textit{ECG5000} and \textit{MedicalImages} confirm that the overall model performance is adequate for use in subsequent experiments.

\begin{figure}[!t]
\centering
\includegraphics[width=\linewidth]{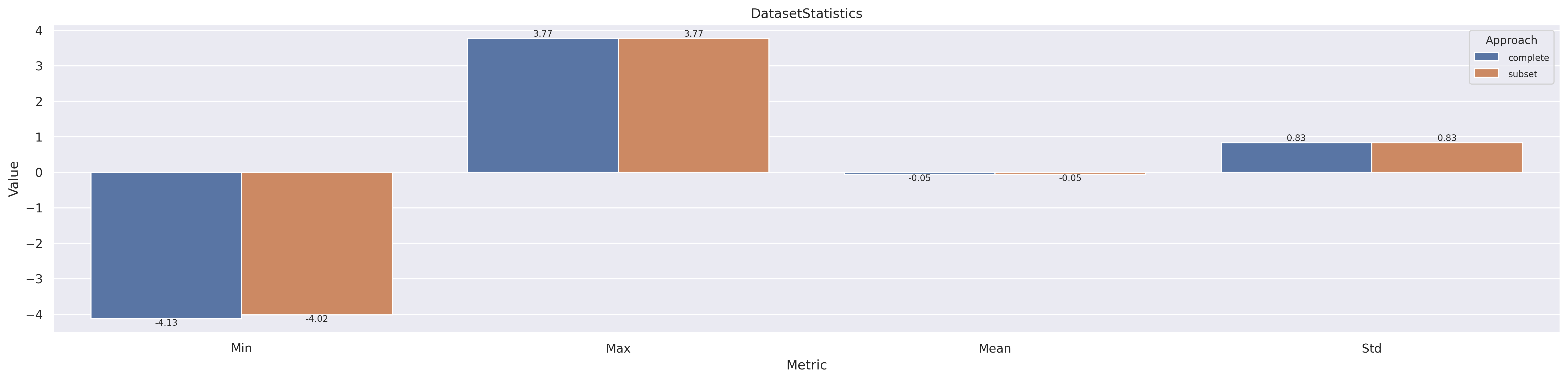}
\caption{\textbf{Dataset Statistics:} Shows different statistical values computed for the full dataset and its subset the \textit{CharacterTrajectories} dataset highlighting that there is no statistical evidence indicating that the subset differs from the full dataset.}
\label{fig:metric_sample_justification}
\end{figure}

\begin{figure}[!t]
	\centering
	\includegraphics[width=\linewidth]{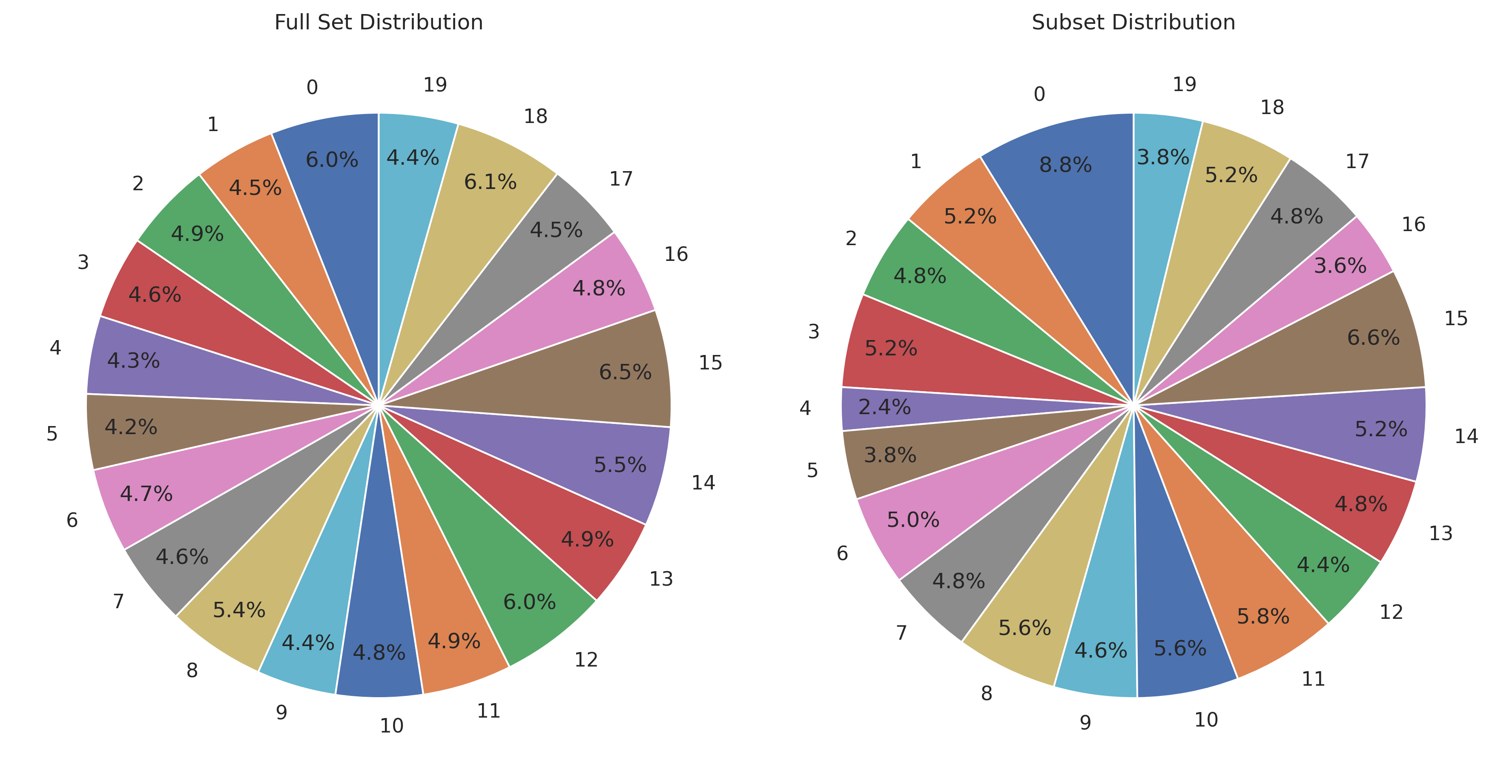}
	\caption{\textbf{Class Distribution:} Shows the class distribution for the full \textit{CharacterTrajectories} dataset and its subset, highlighting that the subset maintains same distribution as the full dataset.}
	\label{fig:distribution}
\end{figure}

For the remaining experiments, it should be noted that the evaluation was performed on a set of $500$ randomly selected samples from each dataset to reduce computational effort. These subsets maintain the same distribution as the original datasets while enabling the computation of metrics that are very computationally expensive and infeasible on the full datasets. However, using $500$ samples per dataset still provides results comparable to those obtained with the complete datasets. Figure~\ref{fig:metric_sample_justification} exemplarily shows the differences between the mean and other statistics of the full \textit{CharacterTrajectories} dataset and its subset. Similarly, Figure~\ref{fig:distribution} visualizes the class distributions in the dataset. Furthermore, to ensure that the samples are neither more nor less complex than the full dataset, the model performance was evaluated on the subset, achieving an accuracy of $0.9600$, a macro F1 score of $0.9543$, and a micro F1 score of $0.9586$ — all close to the original performance scores shown in Table~\ref{tab:accuracy}. The same evaluation was conducted for the remaining datasets, and there was no statistical evidence that any of the subsets deviated significantly in distribution or represented a substantially easier or more difficult set of data samples.

Regarding the hyperparameters, the complete complex FFT was used, as early experiments demonstrated that it outperformses the use of only the magnitude spectrum. Furthermore, the window size for the occlusion — both in the feature space and the frequency space — was set to five, as this value performed well across all evaluated datasets. However, it should be noted that adjusting this value to better suit each dataset individually could improve both traditional and frequency-based attribution approaches. No additional frequency binning was performed, as it was already addressed by the window-based occlusion approach.

\subsection{Overview of statistics across datasets}

\subsubsection{Area under the Curve Using Different Attribution Approaches}

\begin{figure}[!t]
\centering
\includegraphics[width=\linewidth]{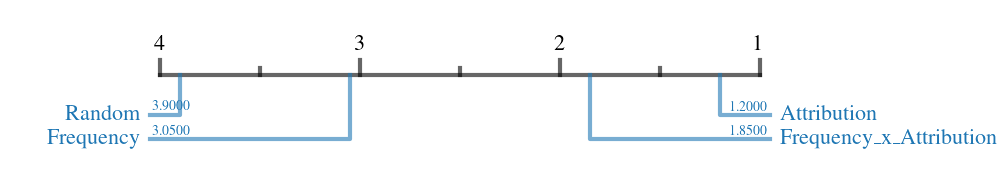}
\caption{\textbf{Area under the curve:} Shows the area under the curve (AUPC) for a deletion test comparing frequency attribution with traditional attribution. Frequency\_x\_Attribution refers to first optimizing the input using frequency attribution and then applying traditional attribution on top. Across all datasets, random attribution exhibited the worst performance.}
\label{fig:cdd_auc}
\end{figure}

To provide a general overview of how frequency attribution performs across all datasets, several well-known metrics were employed. In the first experiment, the area under the performance curve (AUPC) of a deletion test was computed for each dataset. During a deletion test, features are sequentially removed from the signal, which causes the reclassification performance to decrease. The \textit{AUPC} measures the area under the performance curve from the original input to the completely empty input. This relative metric was used to compare random point deletion against traditional attribution, frequency attribution, and their combination. Across all experiments, occlusion was employed as the attribution method for both traditional and frequency attribution processes. The combined approach involves first applying frequency attribution to remove irrelevant frequencies and then applying traditional attribution to highlight important regions. Intuitively, this measurement favors traditional attribution, since individual data points can be precisely adjusted to produce better \textit{AUPC} scores. In contrast, frequency attribution only allows removal of frequencies without adding new ones, which means multiple points are adjusted simultaneously. Figure~\ref{fig:cdd_auc} shows the results for all tested datasets, demonstrating that all attribution approaches significantly outperform random data point deletion. However, this metric primarily provides a ranking of the approaches. For detailed insights into performance gaps between approaches, readers are referred to the dataset-specific analysis, particularly of the \textit{CharacterTrajectories} dataset. Nevertheless, the results indicate that combining traditional and frequency attribution yields better reclassification performance than using frequency attribution alone.

\begin{figure}[!t]
\centering
\includegraphics[width=\linewidth]{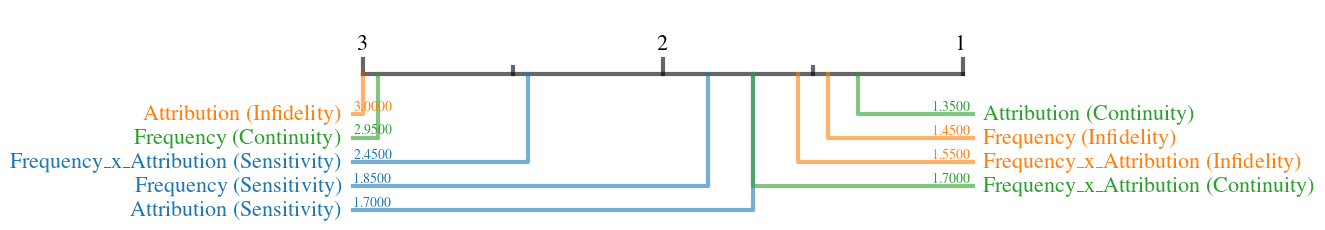}
\caption{\textbf{Infidelity, Sensitivity, and Continuity:} Shows the \textit{Sensitivity}, \textit{Infidelity}, and \textit{Continuity} across all evaluated datasets. The results indicate that the ranking varies significantly depending on the metric; for example, frequency attribution achieves the best performance for \textit{Infidelity} but performs poorly in terms of \textit{Continuity} and \textit{Sensitivity}.}
\label{fig:cdd_sens_inf_cont}
\end{figure}

\subsubsection{Infidelity, Sensitivity, and Continuity of Explanation}
In addition to the \textit{AUPC}, \textit{Infidelity}, \textit{Sensitivity}, and \textit{Continuity} were computed for all datasets. \textit{Infidelity}~\cite{yeh2019fidelity} measures the degree to which the predictor function changes when the explanation is altered. In contrast, \textit{Sensitivity}~\cite{yeh2019fidelity} measures the change in the explanation when the input is slightly perturbed. Thus, \textit{Sensitivity} requires recomputing the attribution, making it computationally more expensive, whereas \textit{Infidelity} does not necessitate additional attribution computation. \textit{Continuity} measures the absolute distance between adjacent time steps in a explanation and thus serves as an indicator of cognitive load. A higher \textit{Continuity} value indicates a noisier signal, making it harder to interpret. Conversely, a lower \textit{Continuity} produces a smoother signal that is easier to understand, as it lacks irrelevant spikes~\cite{ancona2019gradient}. Figure~\ref{fig:cdd_sens_inf_cont} shows the rankings of all approaches for each of these metrics. The results reveal that for \textit{Infidelity}, frequency-based approaches significantly outperform traditional attribution. Conversely, \textit{Sensitivity} ranks the approaches in the reverse order. This occurs because frequency-based approaches exhibit larger changes when the input is perturbed, as perturbations introduce new frequencies that appear in the attribution and are subsequently removed during signal optimization if irrelevant. Finally, \textit{Continuity} results show that traditional attribution produces smoother explanations than frequency-based attribution, since it can precisely highlight specific points while assigning low values to most data points. Frequency attribution lacks this precision, making direct comparisons challenging. However, the combined frequency and traditional attribution approach achieves better \textit{Continuity} than frequency attribution alone.

In summary, evaluating frequency attribution requires additional consideration, as it is not directly comparable to traditional attribution using existing metrics and addresses a slightly different problem. The subsequent experiments demonstrate that frequency attribution excels at filtering signals and focusing on important components, while applying traditional attribution to frequency-optimized signals provides valuable insights into signal importance. Given the substantial computational demands and similar results across datasets to those observed with the \textit{CharacterTrajectories} dataset, subsequent experiments focus exclusively on this dataset for detailed analysis.

\subsection{Visual enhancement}
Time series are inherently complex and may contain information unrelated to the classification task. However, filtering signals to preserve their overall shape while reducing irrelevant information remains challenging. Frequency attribution addresses this challenge by assigning higher attribution scores to frequencies contributing significantly to the decision and lower values to those with only marginal influence on the output. This enables optimization of the input signal to retain only relevant frequencies, facilitating easier interpretation and enabling the application of traditional attribution methods thereafter. The following visualizations demonstrate the benefits of frequency attribution. Particularly in 2D space, it becomes evident that the optimized input preserves the overall shape while removing irrelevant signal components to produce smoother representations.

\subsubsection{Optimized Input Samples in Time Series Space}
To visualize the effects of the different attribution approaches in time series space, the attribution maps are directly applied to the input signal. For frequency attribution, the signal is optimized based on frequency importance. For the combined approach, the signal is first optimized using frequency attribution, followed by traditional attribution. Furthermore, for better comparison, all signals were normalized to maintain their original mean and standard deviation.

\begin{figure}[!t]
	\centering
	\includegraphics[width=\linewidth]{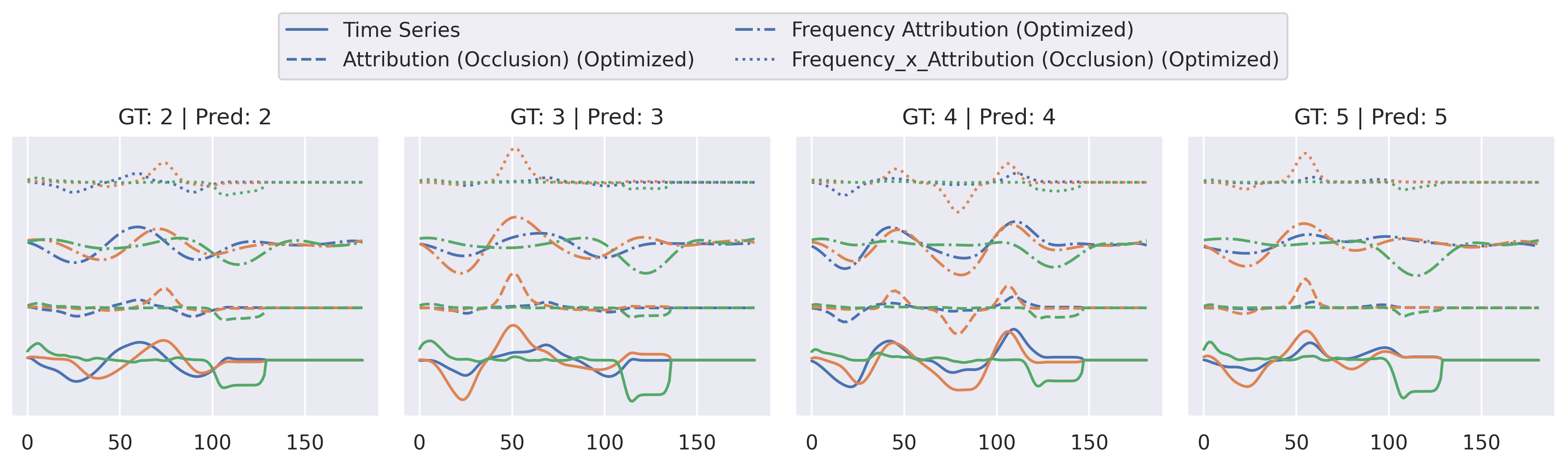}
	\caption{\textbf{Visual comparison of the input in time series space:} Shows the comparison of traditional attribution, frequency attribution, and their combination applied to different character trajectory samples. The frequency-optimized signal exhibits a smoother appearance while preserving the overall shape of the original signal.}
	\label{fig:char_compare_attr}
\end{figure}

Figure~\ref{fig:char_compare_attr} presents the attributed signals for selected samples from the \textit{CharacterTrajectories} dataset. The bottom row shows the complete character, including all information. The frequency attribution produces a smoother version of the original input signal, facilitating interpretation. However, it does not highlight specific highly relevant data points, so traditional occlusion can be applied on top. This is illustrated in the figure, where the overall shapes of the traditional attribution and the combined approach appear very similar. These similar results provide evidence that frequency attribution preserves the essential attribution patterns but reduces irrelevant information that would otherwise require interpretation by the user. Unless stated otherwise, occlusion was used as the attribution method for these experiments. However, additional experiments with other XAI approaches were conducted to compare their performance.

\subsubsection{Optimized Input Samples in 2D Space}

\begin{figure}[!t]
	\centering
	\includegraphics[width=\linewidth]{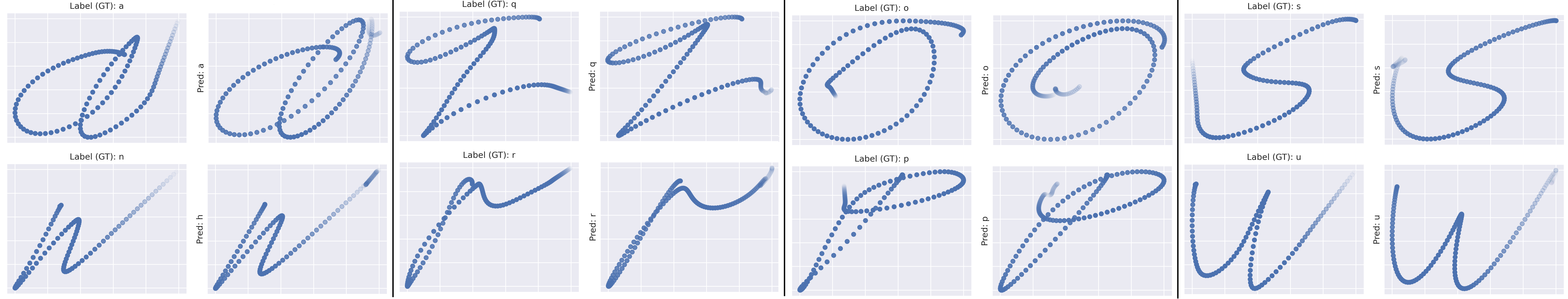}
	\caption{\textbf{Visualization of optimized input in 2D space:} Shows the optimized input for several characters. The left column shows the original input, while the right column displays the frequency-optimized version. The overall shape is preserved while irrelevant parts are removed, resulting in a smoother appearance.}
	\label{fig:char_optimized_rep}
\end{figure}

To demonstrate the impact of signal optimization on the time series signal, Figure~\ref{fig:char_optimized_rep} shows several characters alongside their frequency-optimized versions. Notably, deleting irrelevant frequencies results in only minor changes to the 2D characters. Moreover, for some characters, the visual quality of the 2D representation improves, resulting in smoother appearances. From an explainability perspective, this demonstrates that frequency-optimized versions reduce the cognitive effort required to understand the time series signal and eliminate components that could lead to confusion or misinterpretation, while preserving essential information. Additionally, these modified versions could be used to train or fine-tune models, as they exclude noise and other data points that degrade sample quality.

\subsubsection{Effect of Frequency Attribution on Input}
\begin{figure}[!t]
\centering
\includegraphics[width=\linewidth]{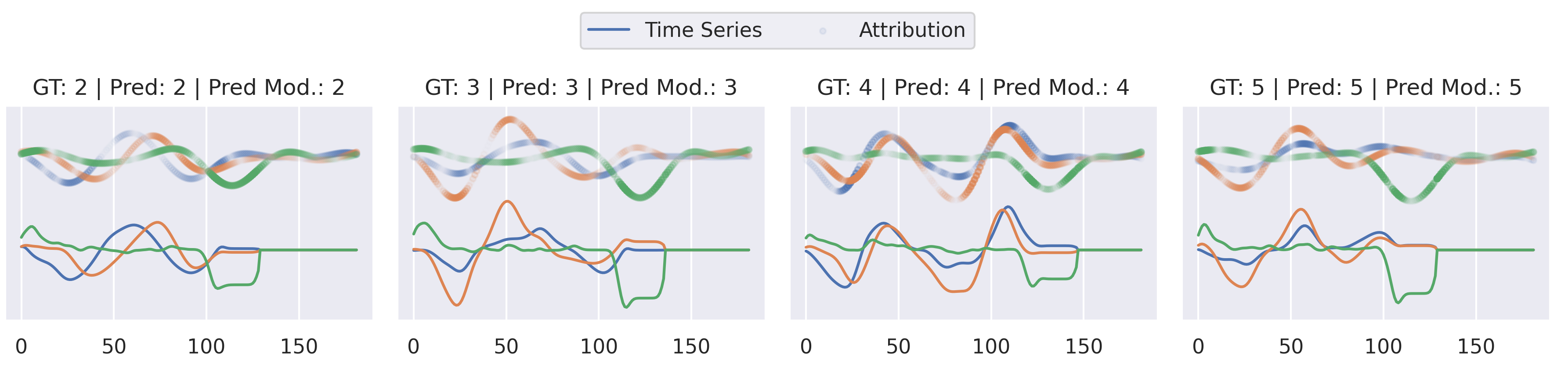}
\caption{\textbf{Signal changes using frequency attribution:} Shows the affected parts of the signal and how frequency attribution adjusts these regions. The color intensity corresponds to the magnitude of signal changes compared to the original input data. The shape corresponds to the optimized signal derived from both the input and the attribution.}
\label{fig:char_freq_attr}
\end{figure}

As the frequency attribution is performed in frequency space, it affects the entire signal. Therefore, occluding a specific frequency can result in substantial changes. To better understand these differences in the input space and precisely identify the affected regions, Figure~\ref{fig:char_freq_attr} shows both the optimized signal and the affected signal components. The figure visualizes the difference between the original and optimized signals as color intensity, with the shape corresponding to the optimized signal. It is evident that nearly the entire signal is adjusted and that small, irrelevant spikes are removed. This demonstrates that frequency attribution effectively cleans the signal.

A compelling example is the green signal, which corresponds to pen pressure. The noise and spikes in the pressure do not contribute to the classification task and were successfully removed by the frequency-optimized signal. Furthermore, it is highly unlikely that the noise in the remaining two signals contributes to the classification task, allowing their removal without losing relevant information. However, this may differ if the task involves writer classification rather than character classification, in which case the network would be trained differently, and the frequency attribution would adapt accordingly.

\subsubsection{Effect of Frequency Attribution on Attribution}
\begin{figure}[!t]
	\centering
	\subfloat[Attribution map (Attribution)]{
		\includegraphics[width=.9\linewidth]{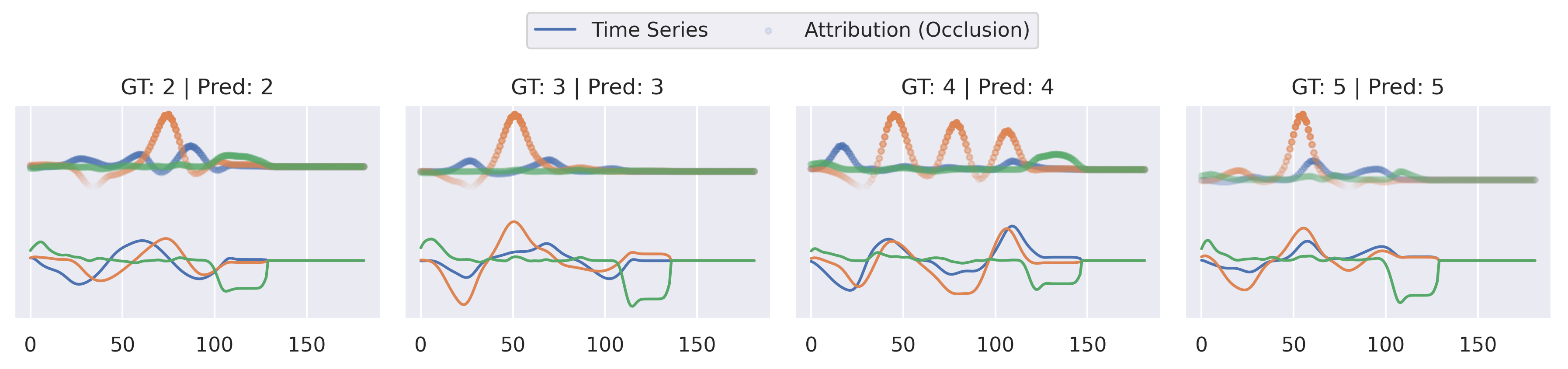}
		\label{fig:char_rep_normal_attr}
	}
	\hfil
	\subfloat[Attribution map (Frequency attribution x attribution)]{
		\includegraphics[width=.9\linewidth]{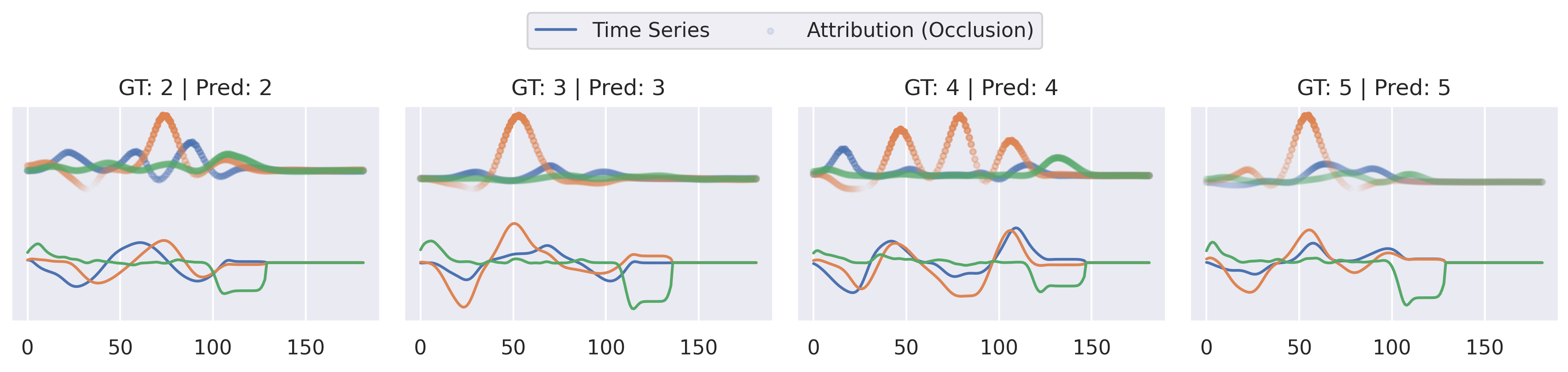}
		\label{fig:char_rep_freq_x_normal_attr}
	}
	\caption{\textbf{Visual comparison of attribution and frequency attribution x attribution:} Shows the comparison of traditional attribution with the combined frequency and traditional attribution approaches applied to different samples from the \textit{CharacterTrajectories} dataset. The results indicate that the overall shapes remain similar.}
	\label{fig:char_rep}
\end{figure}

In the next experiment, the impact on attribution was visualized. Figure~\ref{fig:char_rep_normal_attr} shows the attribution maps for different characters, while Figure~\ref{fig:char_rep_freq_x_normal_attr} shows the corresponding attributions when the signal is first optimized using frequency attribution, as shown in Figure~\ref{fig:char_optimized_rep}. Visually, the attributions appear highly similar. This further emphasizes the value of frequency attribution as a preprocessing step to clean the signal before applying traditional attribution on top. Although the attributions are very similar, interpreting the attribution on the optimized signal is much easier, as irrelevant parts that could lead to confusion have already been removed. For a visualization of the differences between the original and optimized signals, the reader is referred to Figure~\ref{fig:char_optimized_rep}.

\subsection{Statistical Evaluation}
The below statistical evaluation focuses on the \textit{CharacterTrajectories} dataset. However, the experiments were also performed on the other datasets and showed similar results, which are omitted here for readability. Additional figures can be found in the appendix.

\subsubsection{Area under the Curve -- Detailed Evaluation}

To compare the performance of frequency-based attribution with traditional attribution, two experiments were conducted. First, random frequency deletion was compared with the frequency deletion sequence produced by frequency attribution. Second, data point deletion in the input space was evaluated based on attribution ranking. For the first approach, frequencies were removed based on frequency attribution, and the model performance was measured using the modified samples.

\begin{figure}[!t]
\centering
\subfloat[Deletion in frequency space]{
\includegraphics[width=.48\linewidth]{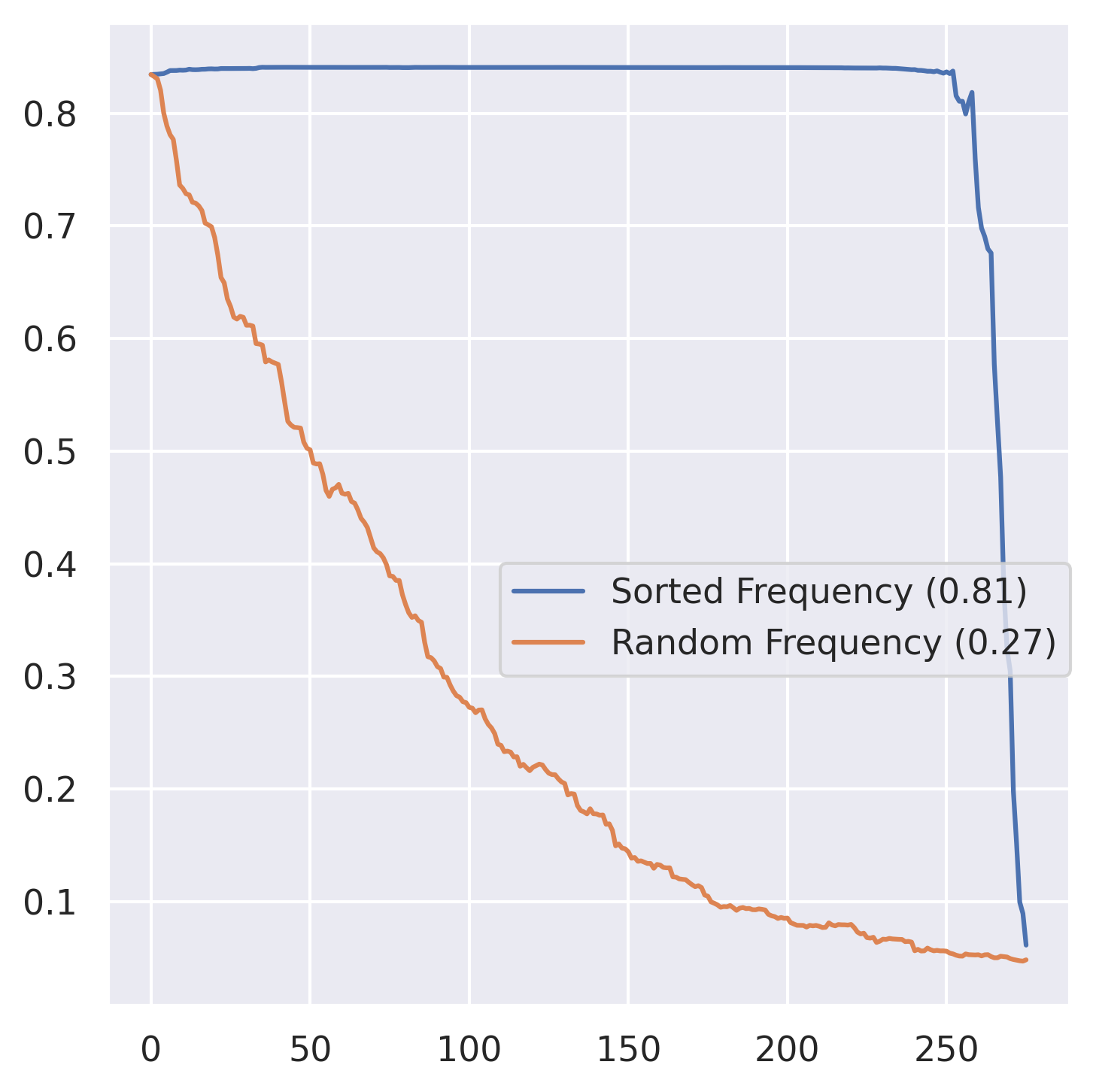}
\label{fig:char_auc}
}
\hfil
\subfloat[Deletion in input space]{
\includegraphics[width=.48\linewidth]{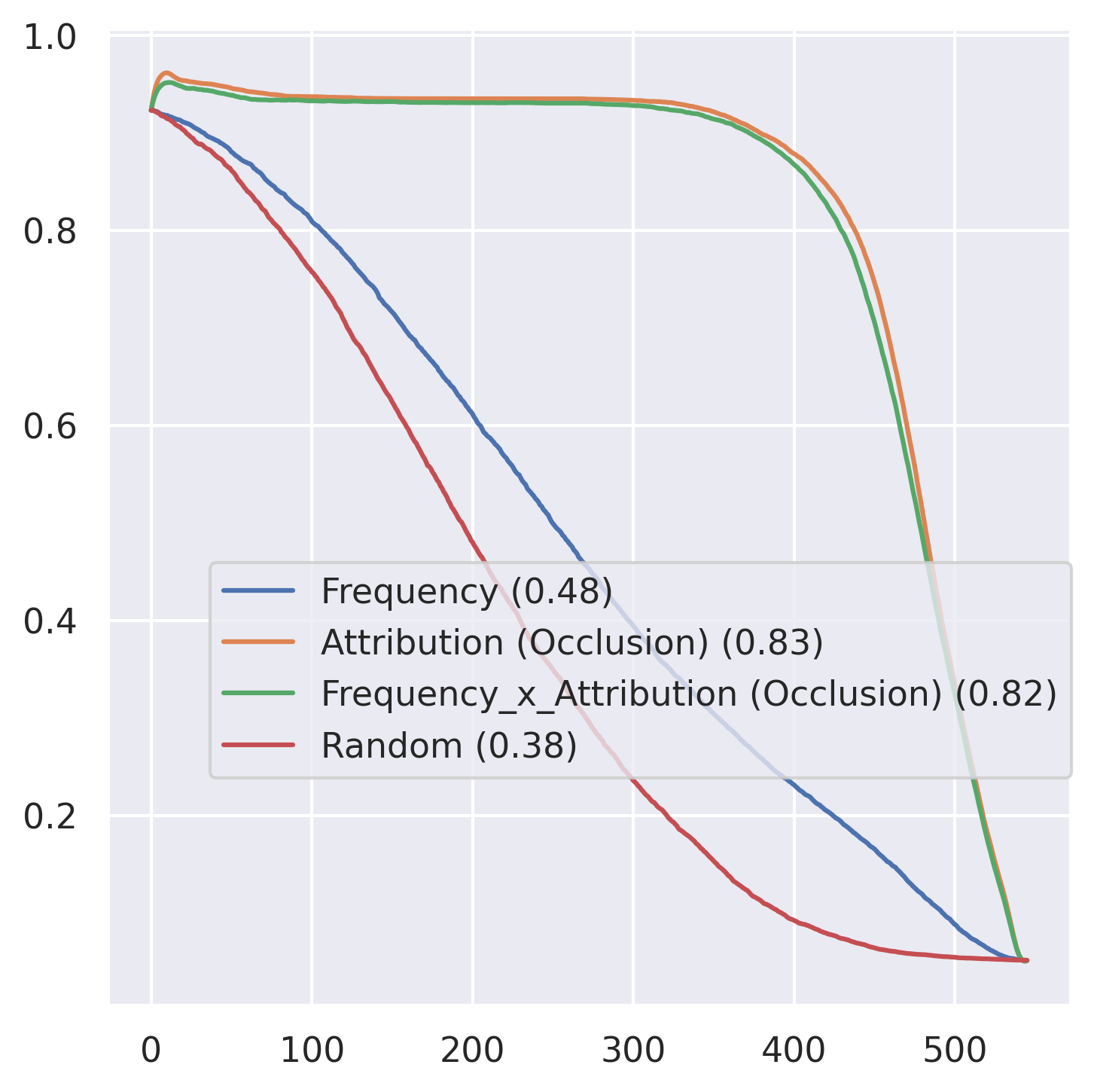}
\label{fig:char_aucs_compare}
}
\caption{\textbf{AUPC deletion test in frequency and input space:} Shows the \textit{AUPC} for the \textit{CharacterTrajectories} dataset comparing removal of random frequencies with frequencies selected by sorted attribution. Additionally, the right panel shows data point removal in the original time series space rather than frequency space. In both experiments, the least important features were removed first.}
\label{fig:char_auc_all}
\end{figure}

Figure~\ref{fig:char_auc} presents the \textit{AUPC} scores comparing random and frequency-based frequency deletion in frequency space. In this experiment, frequencies were sorted such that the least important ones were removed first, which should not drastically affect overall performance. The figure demonstrates that while deleting random frequencies immediately reduces the prediction accuracy, ranked deletion based on frequency attribution maintains performance longer. Performance begins to drop only after removing nearly 200 frequencies, highlighting that many frequencies do not contribute to the classification task. In contrast, deleting a single important frequency significantly reduces performance, as shown by random frequency deletion.

\begin{figure}[!t]
\centering
\includegraphics[width=.94\linewidth]{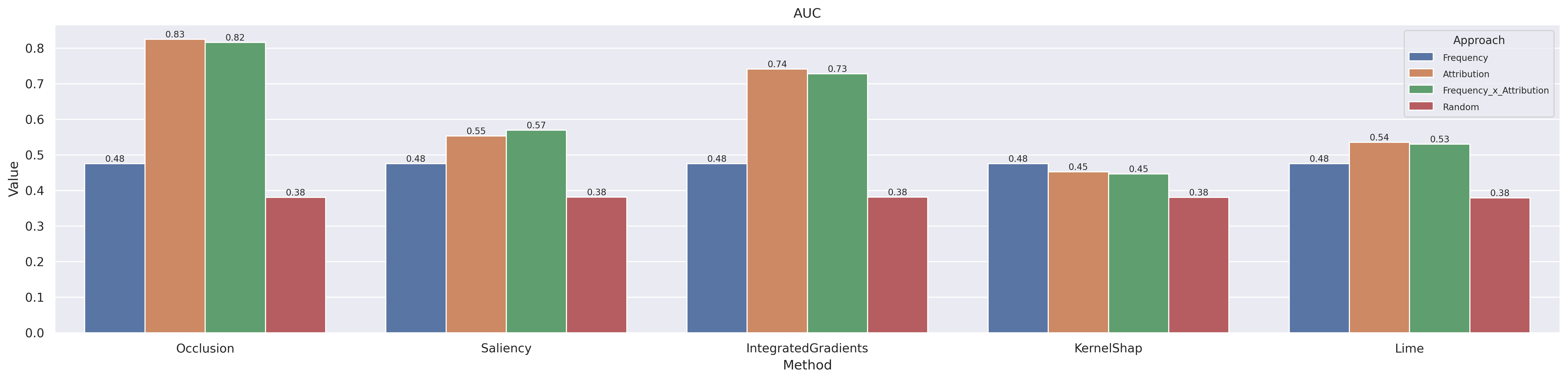}
\caption{\textbf{AUPC different XAI apporaches:} Shows the \textit{AUPC} scores for different XAI approaches on the \textit{CharacterTrajectories} dataset. Higher scores indicate better performance. Traditional attribution and the combination of frequency attribution with traditional attribution exhibit the best performance.}
\label{fig:char_auc_func_bars}
\end{figure}

Figure~\ref{fig:char_aucs_compare} presents deletion results performed in the input space, enabling comparison between frequency-based and traditional approaches. The results indicate that frequency attribution outperforms random deletion but significantly underperforms traditional attribution. However, combining both approaches achieves performance comparable to traditional attribution alone. This occurs because frequency attribution cleans the signal, thereby enhancing traditional attribution performance when applied to the optimized signal. Additionally, the combined approach benefits from a cleaner, more interpretable signal.

In addition to these results, \textit{AUPC} scores were also computed for various traditional attribution approaches, including \textit{Saliency}~\cite{simonyan2013deep}, \textit{Integrated Gradients}~\cite{sundararajan2017axiomatic}, \textit{KernelSHAP}~\cite{lundberg2017unified}, and \textit{LIME}~\cite{ribeiro2016should}. Figure~\ref{fig:char_auc_func_bars} visualizes these methods alongside occlusion. For frequency attribution, masking in the frequency domain was consistently computed using occlusion, whereas for combined frequency and traditional attribution, the traditional method varied. All methods outperformed the random baseline. \textit{Occlusion} achieved the highest \textit{AUPC} of $0.8$, compared to $0.74$ for the second-best method, Integrated Gradients. Notably, both traditional attribution and the combined frequency plus traditional attribution exhibited very similar performance. Interestingly, for \textit{KernelSHAP}, the frequency-only approach outperformed the others.

\subsubsection{Continuity, Infidelity, and Sensitivity of Attributions}

\begin{figure}[!t]
\centering
\includegraphics[width=.94\linewidth]{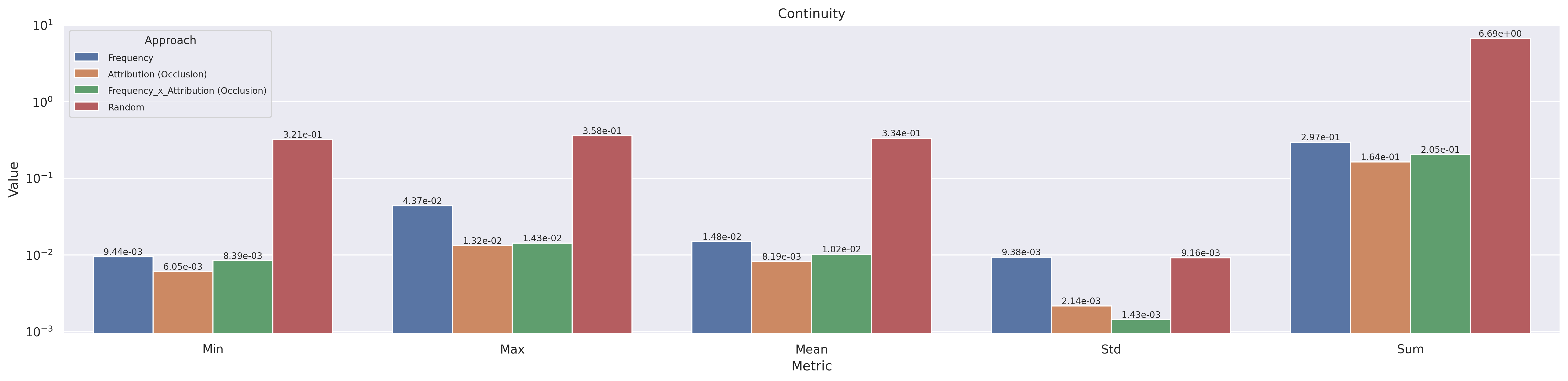}
\caption{\textbf{Continuity:} Shows the \textit{Continuity} results for all approaches. Lower scores indicate better performance. The best results are achieved by combining frequency attribution with traditional attribution.}
\label{fig:char_continuity}
\end{figure}

Besides performance metrics, \textit{Continuity}, \textit{Infidelity}, and \textit{Sensitivity} are commonly used to evaluate attribution quality. Figure~\ref{fig:char_continuity} presents \textit{Continuity} values for all approaches. The results demonstrate that all approaches significantly outperform the random baseline. The figure highlights the superiority of combining traditional and frequency attribution for producing contiguous attribution maps in time series space. Notably, standalone frequency attribution is not directly comparable to other approaches, as it computes attribution as the difference between the original and optimized signals, which inherently produces less continuous results.

\begin{figure}[!t]
\centering
\includegraphics[width=.94\linewidth]{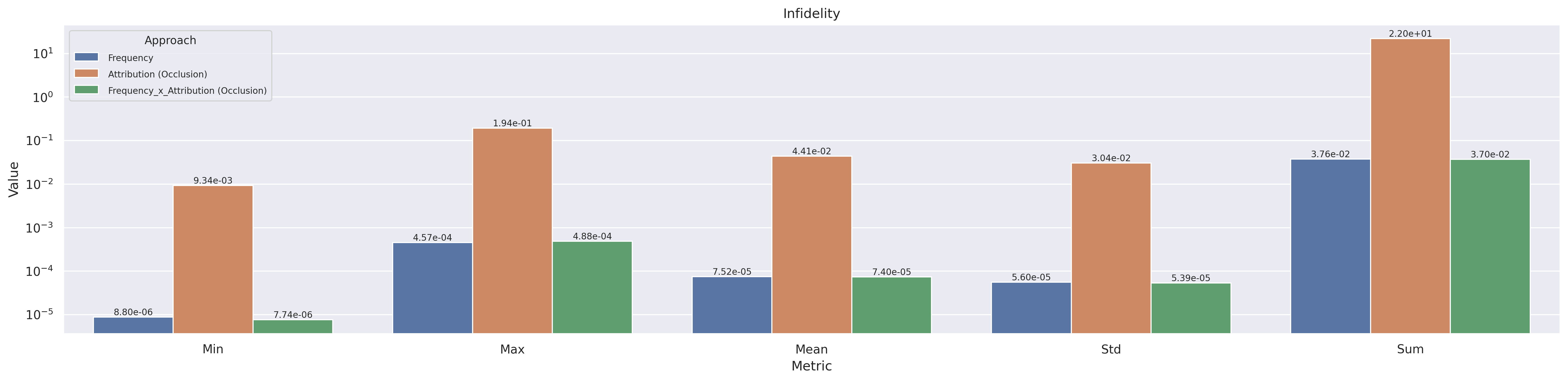}
\caption{\textbf{Infidelity:} Shows the \textit{Infidelity} results for all approaches. Lower scores indicate better performance. The best results are achieved by frequency-based approaches and the combination of traditional and frequency attribution.}
\label{fig:char_infidelity}
\end{figure}

Figure~\ref{fig:char_infidelity} presents the \textit{Infidelity} scores, demonstrating that traditional attribution exhibits higher values. Thus, it is less robust than frequency-based approaches. The results for the combined approach and frequency attribution alone are nearly identical. This confirms that prior signal optimization directly enhances traditional attribution performance, as previously demonstrated.

\begin{figure}[!t]
\centering
\includegraphics[width=.94\linewidth]{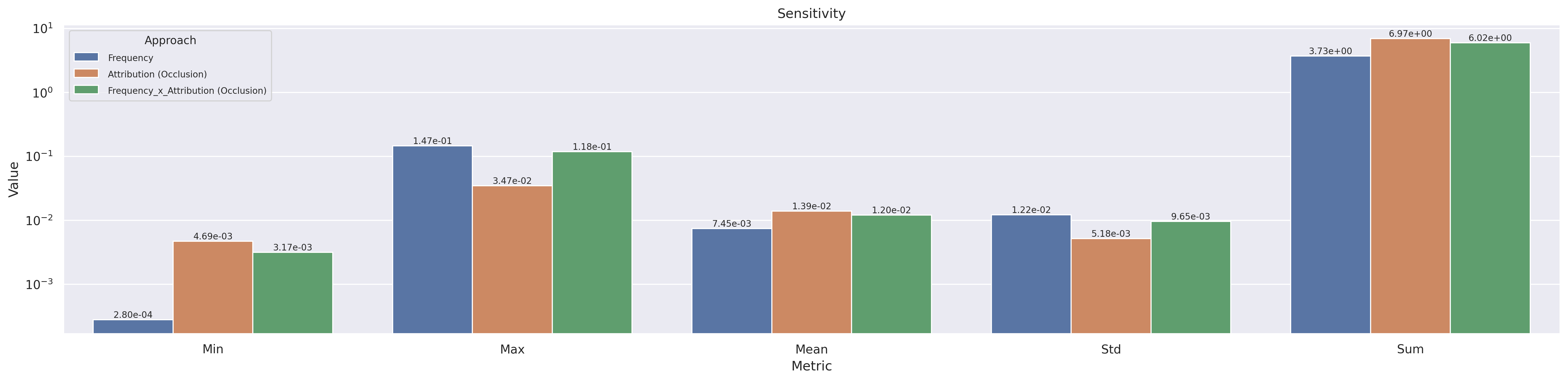}
\caption{\textbf{Sensitivity:} Shows the \textit{Sensitivity} results for all approaches. Lower scores indicate better performance. The best results are achieved by frequency-based approaches.}
\label{fig:char_sensitivity}
\end{figure}

Figure~\ref{fig:char_sensitivity} presents the \textit{Sensitivity} values. These results demonstrate that small changes in the input significantly alter the attributions produced by frequency-based approaches. This \textit{Sensitivity} is not necessarily undesirable, as it indicates that frequency-based approaches detect irrelevant information for classification and adjust the frequency attribution map to remove such changes. Traditional attribution does not exhibit this behavior to the same degree. However, a more detailed analysis is needed for further interpretation of this effect. Additionally, the differences in \textit{Sensitivity} are relatively modest compared to variations observed in the other metrics.

\subsubsection{Mutual Information of Attributed Input}
\begin{figure}[!t]
\centering
\includegraphics[width=.94\linewidth]{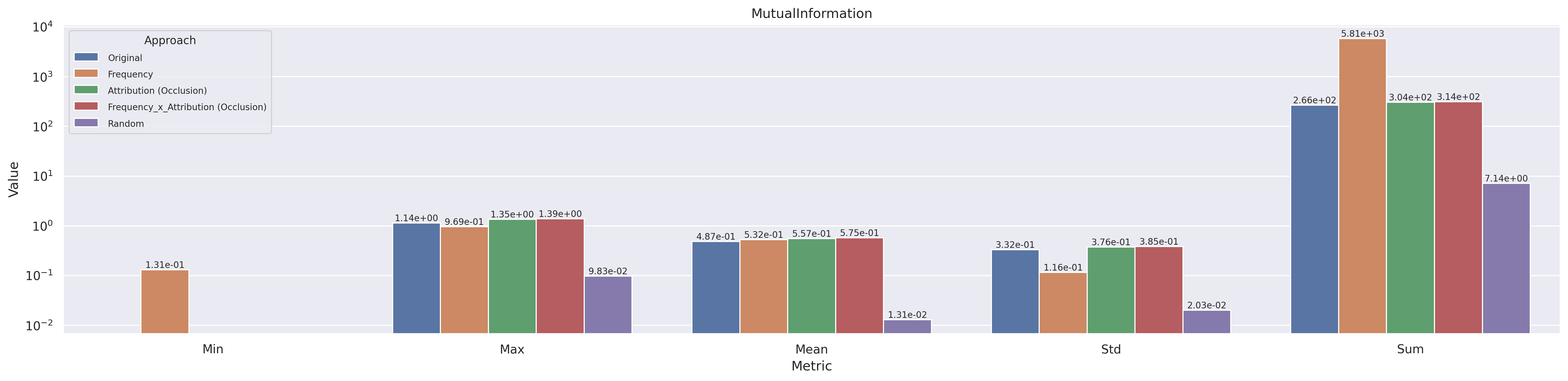}
\caption{\textbf{Mutual Information:} Shows the \textit{Mutual Information} results for all approaches. Higher scores indicate better performance. No clear winner emerges among the methods.}
\label{fig:char_mutual}
\end{figure}

Figure~\ref{fig:char_mutual} presents the \textit{Mutual Information} results for all approaches with respect to the target class. For more information about the \textit{Mutual Information} the reader is referred to Huang et al.\cite{huang2024time}. After applying random attribution, the \textit{Mutual Information} of the signal significantly decreases compared to the original signal. In contrast, both attributed signals and the combination of frequency and traditional attribution slightly increase \textit{Mutual Information}. Notably, standalone frequency attribution achieves a substantially higher \textit{Mutual Information} value, indicating that the smoother signal representation removes class-unrelated information while emphasizing class-relevant components.

\subsubsection{Continuity of Optimized Input Signal}
\begin{figure}[!t]
\centering
\includegraphics[width=.94\linewidth]{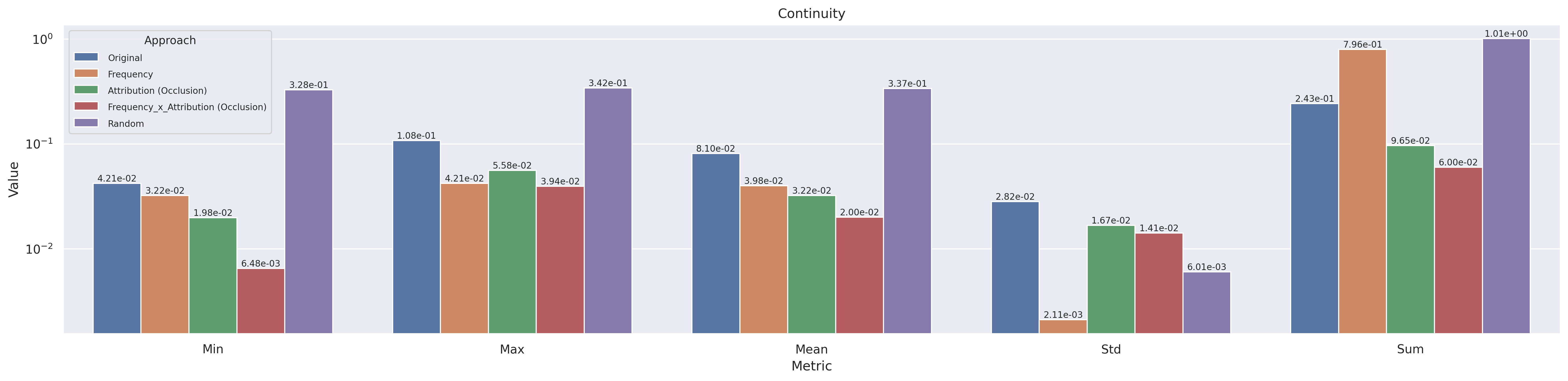}
\caption{\textbf{Continuity Input Signal:} Shows the \textit{Continuity} of the optimized input signals after applying the attribution approaches. Lower scores indicate better performance. The results demonstrate that combining frequency attribution with traditional attribution yields the best continuity.}
\label{fig:char_continuity-input}
\end{figure}

Figure~\ref{fig:char_continuity-input} presents the \textit{Continuity} of the optimized signals. The results show that random attribution produces a noisier input when used to optimize the signal based on data point importance. Interestingly, the frequency-only signal exhibits worse \textit{Continuity} than the original signal, which can be explained by the fact that frequency attribution optimization emphasizes specific regions, resulting in steeper local changes while remaining smooth globally. Both traditional attribution and, particularly, the combination of frequency and traditional attribution achieve the best scores. Additionally, the mean \textit{Continuity} of the frequency-only approach outperforms the original signal.

\subsubsection{Correlation of Attribution and Target Classes}
As a final experiment, the distances between attributions toward different labels were evaluated. Measuring these distances for frequency optimization towards different classes reveals which frequencies characterize specific classes and which classes share common frequency patterns. Three different metrics were employed: first, the l2 distance between the original sample and the optimized version for each label; additionally, Cosine Similarity and Cross-Correlation validated the l2 results. For l2 distance, lower scores indicate better performance, whereas higher scores are preferable for the other two metrics.

\begin{figure}[!t]
\centering
\includegraphics[width=\linewidth]{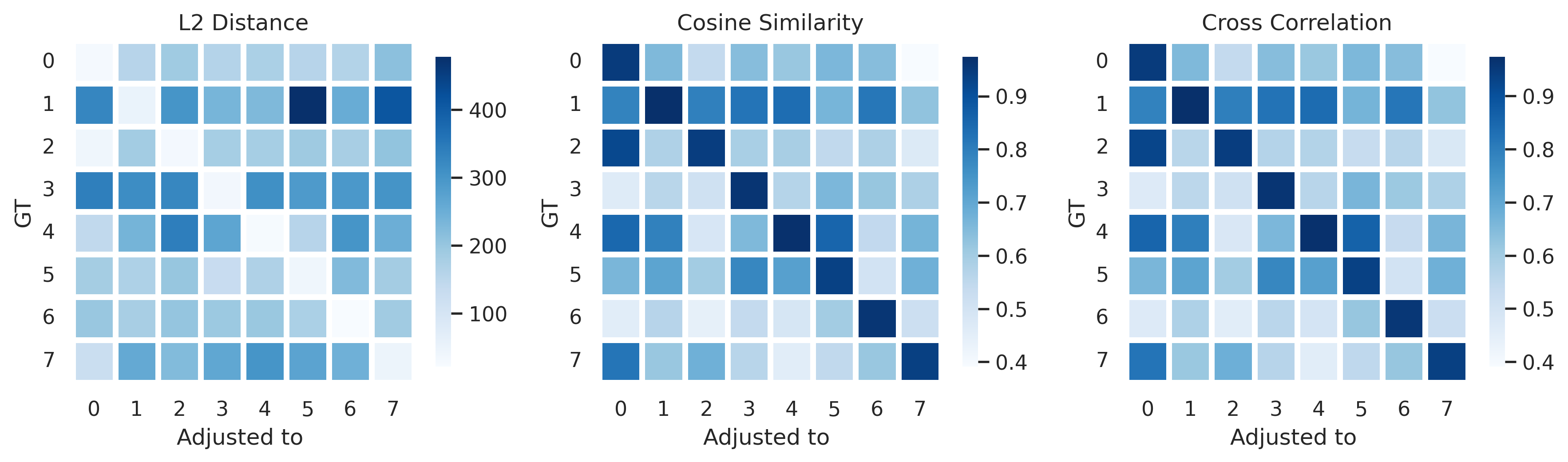}
\caption{\textbf{Distances and similarity:} Shows the distances and similarities between the original sample and frequency-optimized modifications corresponding to other classes. The figure demonstrates that frequency attribution optimization towards the ground truth class preserves high similarity and low distance, while modification toward a different class significantly reduces similarity and increases distance.}
\label{fig:char_distances_similarities}
\end{figure}

Figure~\ref{fig:char_distances_similarities} presents the results for a subset of classes. As expected, the diagonal exhibits the best scores, since optimization toward the correct label remains closer to the original signal than optimization toward other classes. However, certain classes appear very similar in terms of their used frequencies. For example, class zero (character 'a') and class two ('o') share strong similarities, as both letters are written similarly, and their high correlation confirms they utilize common frequencies. Similarly, class one ('n') is clearly distinct from the other two classes, employing different frequencies. This experiment demonstrates that frequency attribution effectively identifies classes sharing frequency patterns and their differences.

\section{Discussion}
\label{sec:discussion}

\subsection{Experiments}
The experiments reveal key insights into frequency-based attribution as a method to optimize and explain time series signals. The \textit{AUPC} results demonstrate that frequency attribution as a standalone approach underperforms traditional attribution — due to fundamental differences in their mechanisms — but combining them significantly improves \textit{AUPC} scores. Similar improvements were observed in \textit{Continuity} and \textit{Sensitivity}, suggesting that hybrid approaches yield superior performance and more comprehensive explanations.

Visual analysis confirms that traditional attribution and the combined approach highlight similar signal regions, with only minor differences in irrelevant areas removed by frequency attribution. Analysis of the \textit{CharacterTrajectories} dataset shows that frequency-based optimization and deletion of irrelevant information preserves character shapes while eliminating confusing noise. Further analysis demonstrates that frequency attribution effectively identifies classes sharing frequency patterns — for example, in the \textit{CharacterTrajectories} dataset, characters 'a' and 'o' share most frequencies.

Additional experiments on \textit{Infidelity}, input continuity, and \textit{Mutual Information} confirm frequency attribution as an effective signal optimization tool that highlights important frequencies. However, existing metrics — designed for traditional attribution — are not perfectly suited for frequency-based methods, suggesting new metrics could better evaluate this approach.

In conclusion, frequency attribution excels at filtering time series data while preserving critical information. Particularly effective against noise, it removes irrelevant components across the entire signal while retaining classification-relevant spikes — a significant advantage over methods that indiscriminately remove peaks. Additionally, it reduces cognitive load by eliminating distracting noise, enabling users to focus on essential explanations. Finally, applying traditional attribution to frequency-optimized signals produces highly effective, clean explanations.

\subsection{Further Existing Techniques}
CENN (Capsule-based Explainable Neural Network)~\cite{zhang2024cenn} and STACN (Spatial-Temporal Attention Convolutional Network)~\cite{zhang2025sparse} are two architectures that capture spatio-temporal features through specialized capsule layers combined with attention mechanisms (CENN) or convolutional layers with temporal attention (STACN). These approaches directly model spatial dependencies and temporal dynamics in input data, primarily focusing on localized features.

Several aspects are shared between these two approaches and the proposed frequency attribution, while others differ significantly. Both architectures extract spatio-temporal representations, but primarily aim to learn patterns from them and use attention mechanisms to highlight important regions. In contrast, frequency attribution directly operates in the frequency domain without extracting or learning features. Instead, it computes the model's sensitivity to specific frequency components rather than identifying important spatial regions.

Although all approaches seek to explain model predictions by identifying important features, CENN and STACN emphasize explicit spatial-temporal patterns and interactions, while frequency attribution reveals the model's sensitivity to periodic or oscillatory components across frequencies. Furthermore, capsule-attention methods like CENN and STACN provide interpretability regarding spatial and temporal features but do not quantify responses to distinct frequency bands. Frequency attribution addresses this gap — unexplored by prior capsule/attention architectures, including the A-LSTM framework for bitcoin-scam detection~\cite{zhao2022attention} — by revealing spectral-domain behavior that cannot be inferred from time-domain attention maps alone.

This is particularly valuable for signals where frequency components encode class distinctions or anomalies overlooked by time-domain spatial networks. Integrating both perspectives — time and frequency domains — enables comprehensive understanding of deep time series models. Finally, frequency attribution remains architecture-agnostic, allowing application to any network.

Zhou et al.~\cite{zhou2022fedformer} developed another approach that incorporates frequency processing by transforming time series into the frequency domain and performing operations directly within that space. The authors demonstrate that this yields improved runtime—owing to the reduced number of frequency bands—and enhanced learning performance, as frequency representations more effectively capture underlying patterns.

These advantages align closely with our findings: \textit{FreqAtt} likewise exhibits linear runtime complexity and superior coverage of long-range patterns compared to traditional time-domain processing methods.

\subsection{Possible Extension}
The proposed method, primarily designed for classification tasks, can also extend to various time series analysis problems, including regression and forecasting. Specifically, frequency attribution could be adapted to time series forecasting by identifying frequency bands most influencing model predictions for future time points, revealing interpretable periodic or seasonal patterns driving forecasts. The difference between the original forecast and the occluded signal serves as an importance measure~\cite{schlegel2021time}.

Similarly, for segmentation tasks, the approach could elucidate how different frequency components contribute to segment boundaries, enabling deeper understanding of underlying signal structures. In anomaly detection, highlighting frequency bands responsible for anomalous behavior would provide valuable interpretability, helping pinpoint spectral signatures triggering anomaly alarms~\cite{kwon2025fft}. These extensions broaden its utility across diverse applications, paving new directions for explainable AI in complex temporal domains.

Additionally, more complex masks could occlude different frequency combinations to analyze their interactions. This requires only modifying the occlusion mask creation function and offers valuable insights into frequency interdependencies. Furthermore, future work could compare frequency attribution against a broader range of attribution methods along with additional faithfulness metrics. This extension would provide deeper validation beyond the scope of this study.

Addressing robust optimization techniques, future research could enable \textit{FreqAtt} to prioritize frequency bands resilient to input perturbations or noise, ensuring stable attribution even under adversarial conditions common in financial or sensor data.

Game-theoretic extensions offer another promising direction, modeling frequency interactions as adversarial games where occluding specific bands reveals robustness against worst-case perturbations.

\subsection{Theoretical Benefits \& Promising Application Domains}
Frequency-domain occlusion offers distinct advantages over traditional time-domain attribution. Intuitively, many time series signals — particularly in biomedical, financial, and environmental monitoring domains — exhibit intrinsic frequency sparsity. Key information often concentrates in specific frequency bands rather than uniformly across time. By occluding frequency components instead of contiguous time segments, the proposed method leverages this sparsity to directly assess meaningful oscillatory patterns that may be obscured in the time domain.

Moreover, frequency attribution aligns with established signal processing principles governing signal energy concentration, where information clusters within narrow spectral bands associated with underlying phenomena. This spectral focus enables the approach to capture interpretable patterns tied to dominant frequencies, complementing time-domain methods that may dilute such effects.

Frequency attribution offers versatile extensions to diverse real-world domains by revealing how specific frequency bands drive model decisions in complex time series. In neuroscience and heart research (e.g., EEG/ECG analysis), it pinpoints oscillatory patterns linked to cognitive states or arrhythmias, enabling precise anomaly detection beyond temporal methods. Financial bubbles and natural gas prediction benefit from uncovering cyclic market regimes or seasonal volatilities, where \textit{FreqAtt} quantifies spectral contributions to crash precursors or demand spikes. In predictive maintenance, machinery vibrations contain frequency signatures of wear or failure; frequency attribution targets these risk indicators more effectively than temporal methods. Environmental monitoring and climate modeling benefit from capturing seasonal effects and oceanic cycles naturally represented in the frequency domain. 

In remote sensing, imaging, and tomography, frequency attribution dissects multi-scale textures in satellite or medical scans, attributing predictions to dominant spatial frequencies for improved invertibility in inverse problems. Networks and coupled economic-behavioral systems gain interpretable insights into propagation dynamics, such as wave-like rumor spread or synchronized oscillations, via frequency-based node importance. Game theory applications model adversarial frequency occlusions to test robustness, while cryptography and information theory leverage it for spectral analysis of encoded signals or entropy patterns.

Overall, frequency attribution provides a complementary — and often superior — lens for interpretability, revealing spectral features inaccessible to conventional approaches and expanding explainable AI's scope.

\section{Conclusion}
\label{sec:conclusion}
Using frequency attribution to optimize time series inputs for subsequent explanations proves highly effective across all datasets. Various established metrics demonstrate that frequency attribution produces representations excluding irrelevant frequencies while preserving those essential for classification. This reduces cognitive effort, enabling focus on truly relevant data components.

Furthermore, combining frequency attribution with traditional attribution outperforms standalone traditional approaches. Although this evaluation focused on occlusion-based attribution, the analysis extends to any attribution method. Additionally, frequency attribution successfully identifies correlated classes based on shared frequency patterns and their correlations.

\backmatter

\bibliography{bibliography}

\iflong
\begin{appendices}
\section{Appendix}
\label{sec:appendix}

\begin{figure}[!ht]
\centering
\includegraphics[width=.94\linewidth]{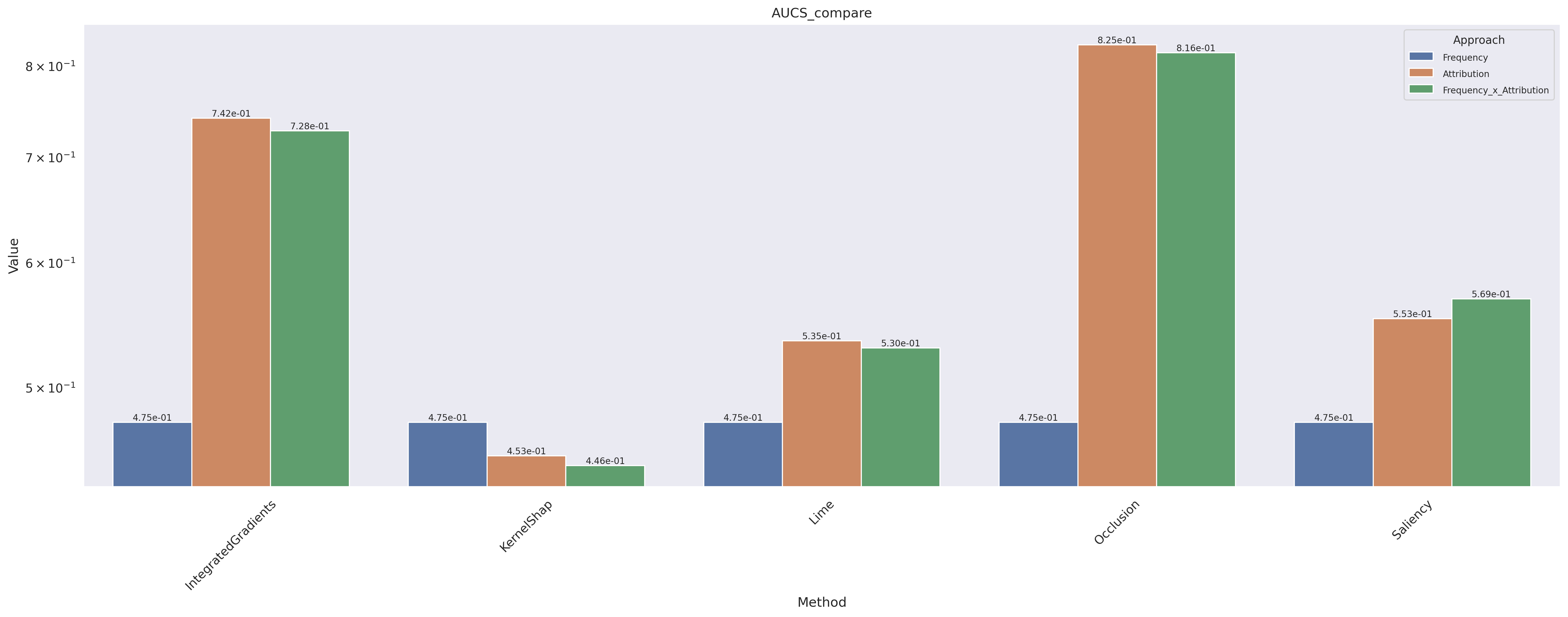}
\caption{\textbf{AUPC Method Comparison:} Shows the \textit{AUPC} sorces of the different attribution approaches for the \textit{CharacterTrajectories} dataset.}
\label{fig:char_AUPCS_all_methods}
\end{figure}

\begin{figure}[!ht]
\centering
\includegraphics[width=.94\linewidth]{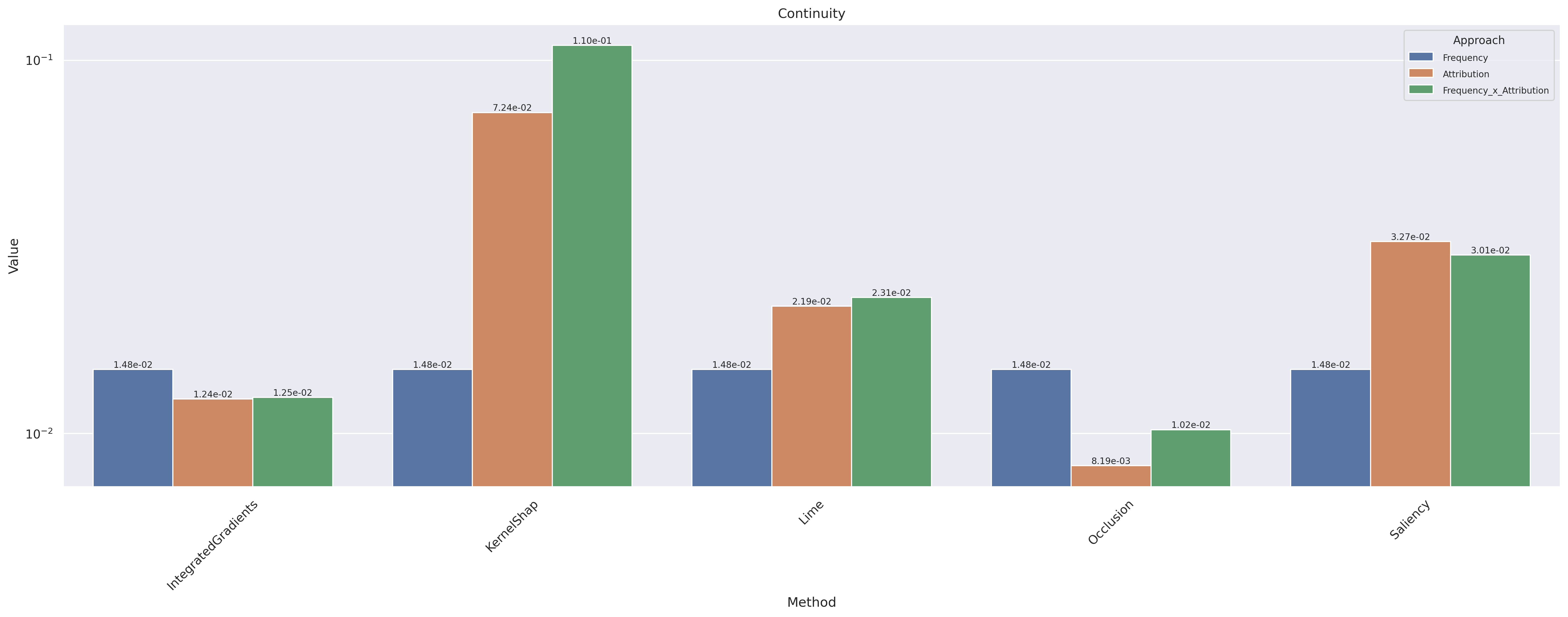}
\caption{\textbf{Continuity Method Comparison:} Shows the \textit{Continuity} of the different attribution approaches for the \textit{CharacterTrajectories} dataset.}
\label{fig:char_Continuity_all_methods}
\end{figure}

\begin{figure}[!ht]
\centering
\includegraphics[width=.94\linewidth]{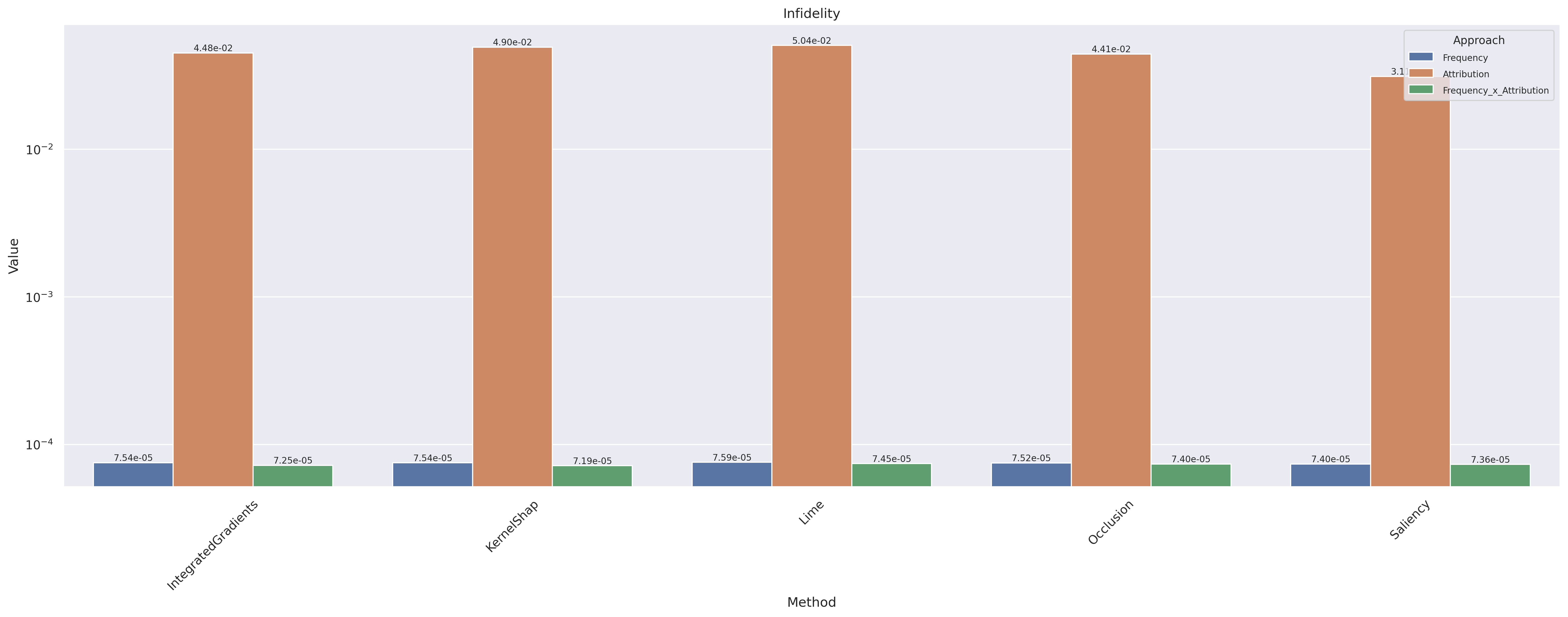}
\caption{\textbf{Infidelity Method Comparison:} Shows the \textit{Infidelity} of the different attribution approaches for the \textit{CharacterTrajectories} dataset.}
\label{fig:char_Infidelity_all_methods}
\end{figure}

\begin{figure}[!ht]
\centering
\includegraphics[width=.94\linewidth]{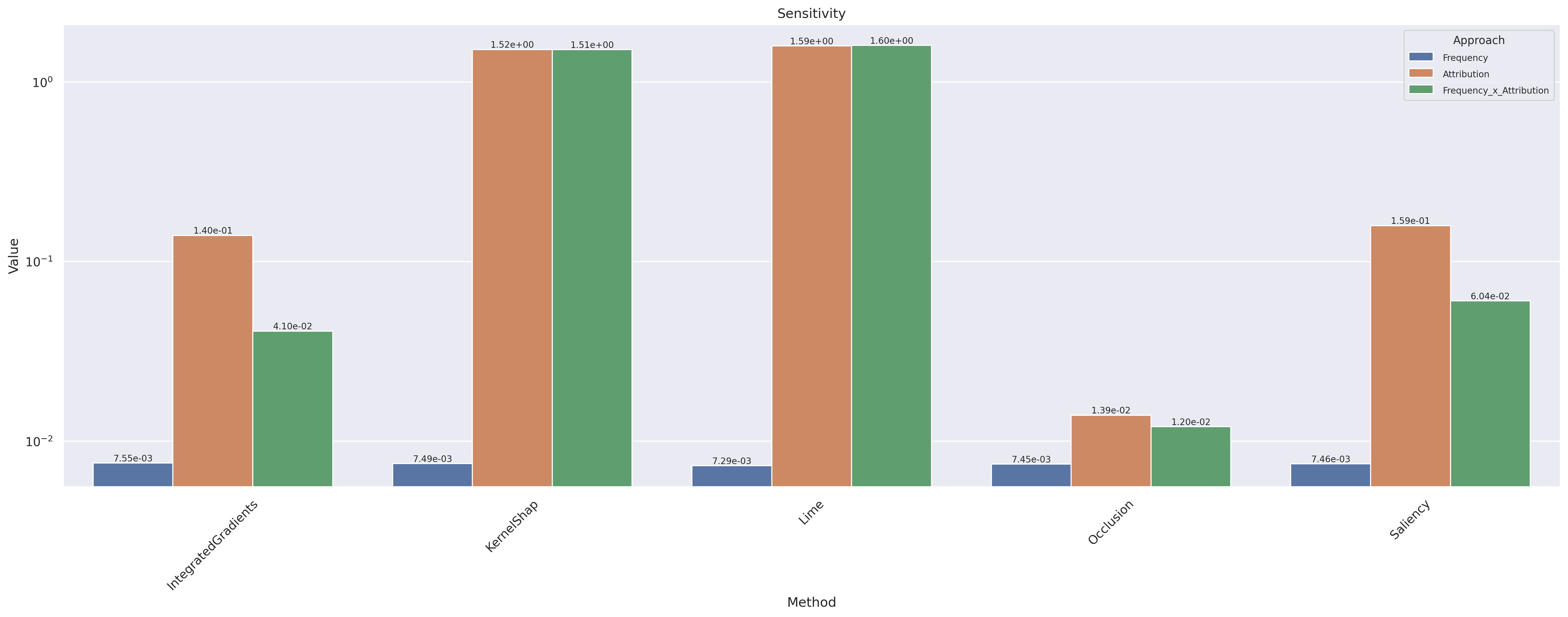}
\caption{\textbf{Sensitivity Method Comparison:} Shows the \textit{Sensitivity} of the different attribution approaches for the \textit{CharacterTrajectories} dataset.}
\label{fig:char_Sensitivity_all_methods}
\end{figure}

\begin{figure}[!ht]
\centering
\includegraphics[width=.94\linewidth]{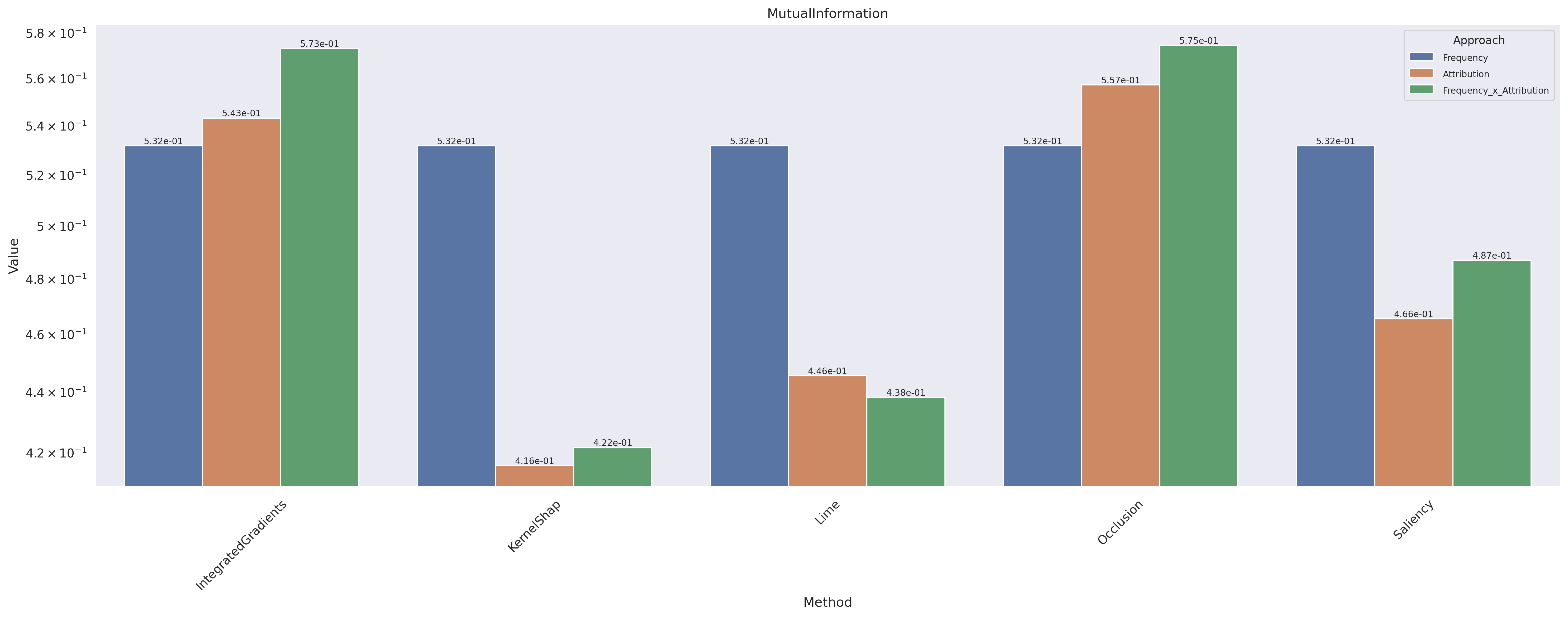}
\caption{\textbf{Mutual Information Method Comparison:} Shows the \textit{Mutual Information} of the different attribution approaches for the \textit{CharacterTrajectories} dataset.}
\label{fig:char_MutualInformation_all_methods}
\end{figure}

\begin{figure}[!ht]
\centering
\includegraphics[width=.94\linewidth]{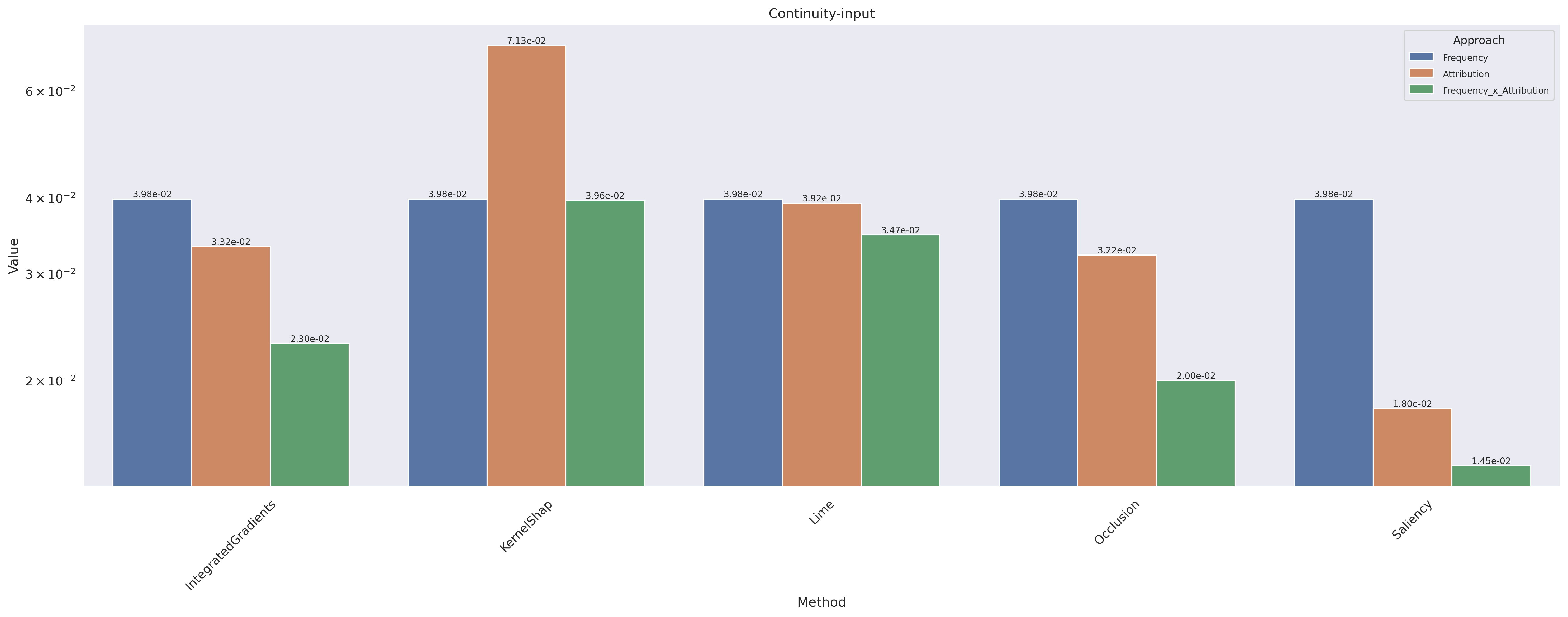}
\caption{\textbf{Input Continuity Method Comparison:} Shows the input continuity of the different attribution approaches for the \textit{CharacterTrajectories} dataset.}
\label{fig:char_Continuity-input_all_methods}
\end{figure}

\end{appendices}
\else
\fi

\end{document}